\documentclass[preprint,10pt]{article}
\usepackage{arxiv}
\usepackage[utf8]{inputenc}
\usepackage[T1]{fontenc}    
\usepackage{hyperref}       
\usepackage{url}            
\usepackage{booktabs}       
\usepackage{amsfonts}       
\usepackage{nicefrac}       
\usepackage{lipsum}         
\usepackage{amsmath}
\usepackage{cleveref}       
\usepackage{amsfonts}
\usepackage{amssymb}
\usepackage{graphicx}
\usepackage{wrapfig}
\usepackage{amsthm}
\usepackage{float}
\usepackage{subfigure}
\usepackage{bm}
\usepackage{caption}
\usepackage{csquotes}
\usepackage[backend=biber]{biblatex} 
\usepackage{caption}
\usepackage{algorithm}
\usepackage{algorithmic}
\usepackage{comment}
\usepackage{xcolor}
\usepackage[english]{babel}
\PassOptionsToPackage{hyphens}{url}\usepackage{hyperref}

\newcommand{\Var}{\operatorname{Var}}
\newcommand{\Expectation}{\operatorname{\mathbb{E}}}

\title{Adaptive importance sampling for Deep Ritz}

\author{{Xiaoliang Wan} \thanks{Department of Mathematics and Center for Computation and Technology, Louisiana State University, Baton Rouge 70803, USA.
Email: xlwan@lsu.edu.}
	\\
	 \And
	 {Tao Zhou} \thanks{ LSEC, Institute of Computational Mathematics and Scientific/Engineering Computing, AMSS, Chinese Academy of Sciences, Beijing, China. Email: tzhou@lsec.cc.ac.cn. }\\
	 \And
	{Yuancheng Zhou } \thanks{ LSEC, Institute of Computational Mathematics and Scientific/Engineering
		Computing, AMSS, Chinese Academy of Sciences, Beijing, China. Email: yczhou@lsec.cc.ac.cn.}
	\\
}

\begin{document}
\maketitle

\begin{abstract}
We introduce an adaptive sampling method for the Deep Ritz method aimed at solving partial differential equations(PDEs). Two deep neural networks are used. One network is employed to approximate the solution of PDEs, while the other one is a deep generative model used to generate new collocation points to refine the training set. The adaptive sampling procedure consists of two main steps. The first step is solving the PDEs using the Deep Ritz method by minimizing an associated variational loss discretized by the collocation points in the training set. The second step involves generating a new training set, which is then used in subsequent computations to further improve the accuracy of the current approximate solution. We treat the integrand in the variational loss as an unnormalized probability density function(PDF) and approximate it using a deep generative model called bounded KRnet. The new samples and their associated PDF values are obtained from the bounded KRnet. With these new samples and their associated PDF values, the variational loss can be approximated more accurately by importance sampling. Compared to the original Deep Ritz method, the proposed adaptive method improves accuracy, especially for problems characterized by low regularity and high dimensionality. We demonstrate the effectiveness of our new method through a series of numerical experiments.
\end{abstract}


\section{Introduction}
Recent years have witnessed fast development in the field of machine learning. There has been a growing interest in utilizing machine learning, particularly deep learning, to address scientific computing problems. Solving partial differential equations (PDEs) is one of the most important tasks in scientific computing. Consequently, many studies on deep learning have been recently conducted to solve problems associated with PDEs. Classical numerical methods such as finite difference method and finite element method face the challenge known as the ``curse of dimensionality'' when dealing with high-dimensional PDEs. In contrast to classical numerical methods, deep learning methods utilize Monte Carlo approximation instead of computational mesh for discretization, which makes them more suitable for solving high-dimensional PDEs. Available deep learning methods can be categorized based on various formulations of PDEs, such as PINNs~\cite{raissi2019physics} and DGM~\cite{sirignano2018dgm} for the strong form of PDEs, WANs~\cite{zang2020weak} for the weak form of PDEs and Deep Ritz~\cite{yu2018deep} for the variational form of PDEs. In this work, our primary focus will be on the Deep Ritz method.

As deep learning methods are meshless approaches, a critical concern is how to select  training points to effectively approximate the integral for the loss functional. The accuracy of the discrete loss functional is pivotal to achieving an accurate approximation of the PDE solution. 
In the original methods of PINNs, DGM, WANs, and Deep Ritz, the loss functionals are approximated using the Monte Carlo method with training points randomly generated from a uniform distribution. However, the uniform sampling strategy is often inefficient if the PDE solution is of low regularity and the situation deteriorates as dimensionality increases. In order to alleviate this problem, some adaptive sampling strategies have been developed to improve the Monte Carlo approximation of the loss functional. However, it's important to note that most adaptive sampling methods are primarily designed for PINNs/DGM, where a change of measure for the integration of the residual does not affect the well-posedness of the formulation. These adaptive sampling strategies include RAR~\cite{lu2021deepxde}, RAD~\cite{wu2023comprehensive}, RAR-G~\cite{wu2023comprehensive}, DAS-G~\cite{tang2023pinns}, FI-PINNs~\cite{gao2023failure}. All of these adaptive sampling strategies provide a certain way to update the current training set by adding some important new training points, which results in a weighted integration of the residual, i.e., the Lebesgue measure is replaced by a new one.  In the Deep Ritz method, the residual term is substituted with a variational term whose minimum value is in general not zero. A straightforward replacement of the Lebesgue measure  
will result in an incorrect variational term such that minimizing the modified loss functional leads to an erroneous solution. As a consequence, the adaptive sampling strategies RAR, RAD, RAR-G, DAS-G, and FI-PINNs cannot be directly applied to the Deep Ritz method. In the Deep Ritz method, it is essential to modify the measure without changing the integral, in other words, importance sampling is a necessity.  
Importance sampling was also considered in DAS-R~\cite{tang2023pinns} for PINNs. 


In this work, we introduce a deep adaptive sampling method for the Deep Ritz method based on importance sampling. Assuming no prior knowledge is available, we intend to use the approximation of the optimal measure for the change of measure required by importance sampling. To approximate the optimal measure induced by the absolute value of the integrand, we need a generic density model. In this work we employ a recently developed deep generative model called bounded KRnet~\cite{zeng2023bounded} to take advantage of its modeling capability for high-dimensional distributions. 
In our adaptive sampling methods, we utilize two deep neural networks: one for approximating the PDE solution and the other one for refining the training set. The approximate PDE solution is attained by minimizing the discretized loss functional, which comprises a variational term and a boundary term. The bounded KRnet defines an invertible transport map from the target distribution on a bounded rectangular domain to a prior (uniform)  distribution on a hypercube $[-1,1]^d$, which is achieved by minimizing the Kullback-Leibler (KL) divergence between the probability density function (PDF) induced by bounded KRnet and the PDF given by the optimal measure.  
We have proposed three adaptive schemes. 
Our basic adaptive scheme is to completely replace the previous training set with new training points generated from the learned bounded KRnet. We also devise two alternative PDF models, which correspond to two different schemes to update the previous training set,  to enhance the performance of the basic adaptive scheme. In all provided adaptive schemes, the updated training set is subsequently employed to further enhance the accuracy of the Deep Ritz method. In summary, the approximate PDE solution and the bounded KRnet will be updated alternately until a certain criterion is satisfied.


The rest of the paper is organized as follows. In the next section, we briefly introduce the Deep Ritz method for the approximation of PDEs. The adaptive sampling method for Deep Ritz is presented in detail in section $\ref{asm for DR}$. Then, in section $\ref{Numerical experiment}$, we demonstrate the performance of our method with numerical experiments. We conclude our work in section $\ref{conclusion}$.

\section{Deep Ritz method for PDEs}
In the Deep Ritz method, our objective is to minimize the following variational problem
$$
\min_{u\in V}I(u)
$$
where $I(u)$,  as a characteristic functional of an elliptic problem defined on a compact domain $\Omega\subset\mathbb{R}^d$ with $d\in\mathbb{N}_+$, takes the form
$$
I(u)=\int_{\Omega} G(u(x))\,dx,
$$
with $G(u)$ being a function of $u(x)$. To deal with the boundary conditions $B(x,u(x))=0$, $x\in \partial \Omega$, we include a penalty term into the objective functional and consider the following optimization problem
\begin{equation}
\min_{u\in V}J(u)
\end{equation}
with
\begin{equation}\label{objective}
J(u)=\int_{\Omega} G(u(x))\,dx+\beta\Vert B(x,u(x))\Vert_{\partial\Omega,2}^2,
\end{equation} 
where $\beta>0$ is a penalty parameter and $\|\cdot\|_{\partial \Omega,2}$ indicates the $L_2$ norm on $\partial \Omega$. 

The Deep Ritz method introduces a neural network $u(\cdot,\theta)$ to approximate the solution $u$ and uses the Monte Carlo method to discretize the integrals in $J(u)$. Then solving the original variational problem becomes minimizing a loss function over the neural network parameters. Let $S_{\Omega}=\left\{ x_i\right\}_{i=1}^{N_v}$ and $S_{\partial \Omega}=\left\{ x_j\right\}_{i=1}^{N_b}$ be two sets of uniformly distributed collocation points on $\Omega$ and $\partial \Omega$ respectively. The discretized optimization problem is  
$$
\min_{\theta}J_N(u(x,\theta))
$$
with
$$
J_N(u(x,\theta))=
\frac{1}{N_v}\sum_{i=1}^{N_v}G(u(x_i,\theta))+\frac{\beta}{N_b}\sum_{j=1}^{N_b}B(x_j,u(x_j,\theta))^2,
$$
which is then solved by stochastic gradient-based methods. Compared to traditional methods, the Deep Ritz method is reasonably simple and less sensitive to the dimensionality of the problem ~\cite{yu2018deep}.

Recently, many adaptive sampling methods have been established to improve the performance of PINNs, most of which are residual-based methods since the objective of PINNs is to minimize the residual on selected collocation points. The Deep Ritz method is not involved with the residual, implying that the residual-based adaptive sampling methods designed for PINNs cannot be directly applied to the Deep Ritz method. In this work, we develop an adaptive sampling method tailored for the Deep Ritz method, with a particular focus on reducing the statistical error in the approximate solution.

\section{Adaptive sampling for Deep Ritz method}\label{asm for DR}
The accuracy of the Deep Ritz method depends on various factors involved in the computation. We are interested in the error induced by discretization, which contains two parts: statistical error and approximation error.  The approximation error comes from the difference between the approximation space and the true space. The statistical error arises from discretizing the loss functional through Monte Carlo approximation. The statistical error depends on the random samples used by the Monte Carlo method while the approximation error relies on the modeling capability of the neural network for the approximation of the PDE solution~\cite{tang2023pinns}. We aim to reduce the statistical error in the Deep Ritz method by seeking a more accurate Monte Carlo approximation of the loss functional, which will be achieved by variance induction through adaptive sampling. 

\subsection{Importance sampling for Deep Ritz method}
For an integrand $G(u)$ with low regularity, the integral can be dominated by the values of $G(u)$ in a small region within $\Omega$. 
In such cases, uniformly increasing the number of training points across the entire domain $\Omega$ may be not effective in reducing the discretization error of the loss functional.  To illustrate this point, we provide the following example. Suppose $G(u(x))=1/\sqrt{2\pi}\exp(-x^2/2\sigma^2)$ on $\Omega =[-1,1]$ and $\sigma\ll 1<1/3$. As the parameter $\sigma$ decreases, the shape of $G(u(x))$ becomes sharper around zero. Using $3\sigma$ principle, the integral can be computed as
$$
\int_{\Omega}G(u(x))\,dx=\int_{\Omega}\frac{1}{\sqrt{2\pi}}\exp(-\frac{x^2}{2\sigma^2})\,dx\approx\sigma.
$$
The Monte Carlo approximation of the integral is
$$
\hat{I}=\frac{1}{N}\sum_{i=1}^{N}\frac{1}{\sqrt{2\pi}}\exp(-\frac{x_i^2}{2\sigma^2})
$$
where $\left\{x_i\right\}$, $i=1,\cdots,N$ is a set of uniformly samples in $\Omega$. The relative error of the approximation is
$$
\frac{(\Var{I})^{\frac{1}{2}}}{\sigma}\approx \frac{1}{\sigma}\sqrt{\frac{\sqrt{2}\Gamma(\frac{5}{4})\sigma}{2\pi N}}=C(\sigma N)^{-1/2}
$$
where $\Gamma(\cdot)$ is the Gamma function and $C$ is a constant. Then, in order to obtain a relative error of $O(1)$, we need around $O(1/\sigma)$ number of samples. This means the uniform samples are not effective for this example. In order to improve the accuracy, more effective samples generated by variance reduction techniques are required. 

One of the most commonly used variance reduction techniques is importance sampling. The idea of importance sampling is to generate a training set from a well-chosen distribution and get an unbiased estimator, 
\begin{equation}\label{importance sampling}
\begin{aligned}
I(u)=\int_{\Omega}G(u(x,\theta))\,dx&=\int_{\Omega}\frac{G(u(x,\theta))}{p(x)}p(x)\,dx\\
&=\Expectation_p\left[\frac{G(u(X,\theta))}{p(X)}\right]\approx\frac{1}{N_v}\sum_{i=1}^{N_v}\frac{G(u(x_i,\theta))}{p(x_i)}
\end{aligned}
\end{equation}
where $\left\{x_i\right\}$, $i=1,\cdots,N_v$ is a set of samples from the probability density function $p(x)$. If the variance of $G(u(X,\theta)/p(X))$ with respect to density function of $p$
\begin{equation} 
\begin{aligned}\label{variance}
\sigma_{p}=\Var_p\left[ \frac{G(u(X,\theta))}{p(X)}\right]&=\Expectation_p\left[\frac{G^2(u(X,\theta))}{p^2(X)}\right]-\left(\Expectation_p\left[\frac{G(u(X,\theta))}{p(X)}\right]\right)^2\\
&=\int_{\Omega}\frac{G^2(u(x,\theta))}{p(x)}\,dx-\left(\int_{\Omega}G(u(x,\theta))\,dx\right)^2
\end{aligned}
\end{equation}
is smaller than that of $G(u(x,\theta))$ with respect to the uniform distribution on $\Omega$, the performance of the Monte Carlo method will be improved. 

From importance sampling \eqref{importance sampling}, if we assume $G(u(x,\theta))\ge 0$ for all $x\in \Omega$, one can find the best choice of the density function $p(x)$ is
$$
p^{\star}(x)=\frac{G(u(x,\theta))}{\mu}
$$
where $\mu=\int_{\Omega}G(u(x,\theta))\,\mathrm{dx}$. We can check this fact by plugging the $p^{\star}$ into equation \eqref{variance}. The variance is zero which means the error is also zero. In practice, $\mu$ is what we need to compute and therefore we do not know $p^{\star}$, however, results suggest variance reduction can be achieved if we can find a $p$ close to $p^{\star}$~\cite{tang2023pinns}.

In the Deep Ritz method, one may encounter the situation when the integrand $G(u(x,\theta))$ is not always non-negative for $x\in \Omega$. For this case, the best choice for the selection of $p(x)$ is
\begin{equation}\label{best_p}
p^{\star}(x)=\frac{|G(u(x,\theta))|}{\mu},
\end{equation}
where $\mu=\int_{\Omega}|G(u(x,\theta))|\,dx$. More specifically, we have 
for any density function $p(x)$
\begin{equation}\label{proof of best p}
\begin{aligned}
\Expectation_{p^{\star}}\left[\frac{G^2(u(X,\theta))}{{p^{\star}}^2(X)}\right]&=\int_{\Omega}\frac{G^2(u(x,\theta))}{p^{\star}(x)}\,dx\\
&=\left(\int_{\Omega}|G(u(x,\theta))|\,dx\right)^2\\
&=\left(\int_{\Omega}\frac{|G(u(x,\theta))|}{p(x)}p(x)\,dx\right)^2\\
&\le \int_{\Omega}\frac{|G(u(x,\theta))|^2}{p^2(x)}p(x)\,dx\cdot\int_{\Omega}p(x)\,dx\\
&=\Expectation_{p}\left[\frac{G^2(u(X,\theta))}{p^2(X)}\right],
\end{aligned}
\end{equation}
where the Cauchy-Schwarz inequality is applied. 
We add $-(\int_{\Omega}G(u(x,\theta))\,\mathrm{dx})^2$ to both side of inequality \eqref{proof of best p} and obtain 
$$
\sigma_{p^{\star}}=\Var_{p^{\star}}\left[ \frac{G(u(X,\theta))}{p^{\star}(X)}\right]\le \Var_p\left[ \frac{G(u(X,\theta))}{p(X)}\right]=\sigma_p.
$$
Also,  we know from the proof
$$
\sigma_{p^{\star}}=\left(\int_{\Omega}|G(u(x,\theta))|\,dx\right)^2-\left(\int_{\Omega}G(u(x,\theta))\,dx\right)^2>0,
$$
if the integrand $G(u(x,\theta))$ contains both positive part and negative part. Then the error of the Monte Carlo approximation satisfies
$$
\Var_p^{\frac{1}{2}}\left[\frac{1}{N_v}\sum_{i=1}^{N_v}\frac{G(u(X_i,\theta))}{p(X_i)}\right]=\frac{\sigma_p}{\sqrt{N_v}}\ge \frac{\sigma_{p^{\star}}}{\sqrt{N_v}}>0.
$$
Unlike the case when the integrand $G(u(x,\theta))$ is non-negative, 
the variance will not reduce to zero but to a positive number. If we want to further reduce the variance after $p$ reaches to $p^{\star}$, we have to increase the number of samples.

\subsection{Bounded KRnet: PDF approximation and sample generation}
From the previous section, we know that $p^{\star}=|G(u(x,\theta))|/\mu$ for a fixed parameter $\theta$ is theoretically the optimal density function for implementing importance sampling in the Deep Ritz method. In order to calculate the estimator in $(\ref{importance sampling})$, we may choose $p$ that is sufficiently close to the $p^{\star}$. However, PDF approximation is a challenging task, especially in high-dimensional spaces. Since the shape of the graph $|G(u(x,\theta))|/\mu$ can be complex in high-dimensional cases, it is hard to use the exponential family of distributions and mixture models to approximate this function. Thus, deep generative models which have stronger representation ability are taken into consideration. In particular, we use a recently developed model, called bounded KRnet~\cite{zeng2023bounded}, to approximate $p^{\star}$ and generate samples as well. Bounded KRnet is a kind of normalizing flow~\cite{papamakarios2021normalizing}. It is an invertible transport map between a prior distribution and the target distribution. The bounded KRnet is defined on a compact domain while the original KRnet is defined on the whole space. This difference makes bounded KRnet a more suitable tool than the original KRnet to approximate $|G(u(x,\theta))|/\mu$ since in most cases, the computation domain of a PDE is compact.

Suppose $\Omega\subset \mathbb{R}^d$ is a compact set in $\mathbb{R}^d$. Let $X\in \Omega$ be a random vector associated with the target PDF $p_{X}(x)$. Let $Z\in \Omega$ be another random vector with a known PDF $p_{Z}(z)$, e.g., PDF of the uniform distribution. Our goal is to find an invertible transport map $f:\Omega\rightarrow \Omega$ from $X$ to $Z$. Once we obtain the map $f$, we can calculate the PDF of $X=f^{-1}(Z)$ using the change of variables 
\begin{equation}\label{change of variables}
p_{X}(x)=p_{Z}(z)|\det \nabla_{x}f|.
\end{equation}
This is an explicit density model for $X$. Using the inverse map $f^{-1}$, we can also easily obtain exact samples of $X$ by $X=f^{-1}(Z)$. The idea of KRnet originates from Knothe-Rosenblatt rearrangement~\cite{santambrogio2015optimal,carlier2010knothe}. It integrates the K-R rearrangement into the definition of the invertible transport map $f$. In practice, the invertible map $f$ is composed of a sequence of simple invertible maps and each map in the sequence is a shallow neural network. In bounded KRnet, these simple invertible map is constructed by a non-linear map from $[-1,1]^d$ to $[-1,1]^d$ while in original KRnet, the map is from $\mathbb{R}^d$ to $\mathbb{R}^d$.

Suppose $f_{\text{bKRnet}}(\cdot,\theta_f)$ is the invertible transport map induced by the bounded KRnet where $\theta_f$ includes the model parameters. According to equation \eqref{change of variables}, we can explicitly write down the PDF with respect to $f_{\text{bKRnet}}$
$$
p_{\text{bKRnet}}(x,\theta_f)=p_Z(f_{\text{bKRnet}}(x,\theta_f))|\det \nabla_{x}f_{\text{bKRnet}}|
$$
The sample of $p_{\text{bKRnet}}(x,\theta_f)$ can be generated through $X=f_{\text{bKRnet}}^{-1}(Z)$ by sampling $Z$. In bounded KRnet, the random vector $Z$ is typically assumed to be a uniform random variable on the cube $[-1,1]^d$. Our aim is to find out the optimal parameters $\theta_f^{\star}$ such that the PDF $p_{\text{bKRnet}}(x,\theta_f^{\star})$ 
is close enough to a target PDF $p$. In order to find out $\theta_f^{\star}$, we minimize the KL divergence from $p$ to $p_{\text{BKRnet}}(x,\theta_f^{\star})$, i.e.,
\begin{equation}\label{KL objective}
\theta_f^{\star}=\mathop{\arg\min}_{\theta_f}D_{\text{KL}}(p\,||\,p_{\text{bKRnet}},(x,\theta_f)).
\end{equation}
where
\begin{equation}\label{KL split}
\begin{aligned}
&D_{\text{KL}}(p\,||\,p_{\text{bKRnet}}(x,\theta_f))\\
=&\int_{\Omega}p(x)\log(p(x))\,{dx}-\int_{\Omega}p(x)\log(p_{\text{bKRnet}}(x,\theta_f)))\,{dx}.
\end{aligned}
\end{equation}
It can be seen the first term in equation \eqref{KL split} is irrelevant to the density model and therefore can be ignored. Hence the minimization problem \eqref{KL objective} is equivalent to minimizing the second term in equation \eqref{KL split} which is the cross entropy $H(p,p_{\text{bKRnet}})$ from $p$ to $p_{\text{bKRnet}}(x,\theta_f^{\star})$. That is
\begin{equation}\label{cross entropy}
\begin{aligned}
\theta_f^{\star}=& \mathop{\arg\min}_{\theta_f}H(p,p_{\text{bKRnet}}).
\end{aligned}
\end{equation}
For the Deep Ritz method, we set $p=p^{\star}$ in equation \eqref{best_p}. We also apply importance sampling to the computation of the cross entropy in problem \eqref{cross entropy} using a prior PDF model $\tilde{p}$, i.e.
\begin{equation}
H(p^{\star},p_{\text{bKRnet}})\approx-\frac{1}{N_v}\sum_{i=1}^{N_v}\frac{|G(u(x_i,\theta))|\log(p_{\text{bKRnet}}(x_i,\theta_f))}{\mu \tilde{p}(x_i)},
\end{equation}
where $\left\{x_i\right\}$, $i=1,\cdots,N_v$ are samples generated from the prior PDF $\tilde{p}$. 
The prior PDF $\tilde{p}$ will be specified in our adaptive sampling procedure presented in the next section.


\newpage

\subsection{Adaptive sampling method}
We now present our adaptive sampling algorithms. We first present a basic algorithm, where all collocation points in the training set will be replaced at each adaptivity iteration in terms of a new density model. Then we propose two less aggressive but more robust strategies, where the training set is partially updated at each adaptivity iteration in terms of a mixture density model.

\subsubsection{Basic adaptive algorithm(replace all training points)}
Let $S_{\Omega}^k=\left\{ x_{\Omega,i}^k,i=1,\cdots,N_v\right\}$ and $S_{\partial\Omega}^k=\left\{ x_{\partial\Omega,j}^k,j=1,\cdots,N_b\right\}$ be two sets of collocation points at the $k$th adaptivity iteration for the Deep Ritz method, which respectively deal with the variational term and the penalty term for the boundary conditions. 
Let $S_{B}^k=\left\{ x_{B,l}^k,l=1,\cdots,N_B\right\}$ be the training points for bounded KRnet at the $k$th adaptivity iteration. In the $(k+1)$th adaptivity iteration, we first solve the following optimization problem from the Deep Ritz method
\begin{equation}\label{A1}
\theta^{k+1}=\mathop{\arg\min}_{\theta}\frac{1}{N_v}\sum_{i=1}^{N_v}\frac{G(u(x_{\Omega,i}^k,\theta))}{p_{\text{bKRnet}}(x_{\Omega,i}^k,\theta_{f}^k)}+\frac{\beta}{N_b}\sum_{j=1}^{N_b}B^2(x_{\partial\Omega,j}^k,u(x_{\partial\Omega,j}^k,\theta))
\end{equation}
where $p_{\text{bKRnet}}(\,\cdot\,,\theta_{f}^k)$ is the PDF given by bounded KRnet when $k>0$, and chosen as a unfirom distribution when $k=0$. 
Correspondingly, $S_{\Omega}^k$ is sampled from a uniform distribution when $k=0$ while it is sampled from $p_{\text{bKRnet}}(\,\cdot\,,\theta_{f}^k)$ when $k>0$. Once we have computed $\theta^{k+1}$, we can update the bounded KRnet by 
\begin{equation}\label{A2}
\theta_f^{k+1}=\mathop{\arg\min}_{\theta_f}-\frac{1}{N_B}\sum_{l=0}^{N_B}\frac{|G(u(x_{B,l}^k,\theta^{k+1}))|\log(p_{\text{bKRnet}}(x_{B,l}^{k},\theta_f))}{\mu^{k+1}p_{\text{bKRnet}}(x_{B,l}^k,\theta_f^{k})}
\end{equation}
where $S_{B}^k$ is sampled from a uniform distribution when $k=0$ while it is sampled from $p_{\text{bKRnet}}(\,\cdot\,,\theta_{f}^k)$ when $k>0$. This way, the training set $S_{B}^k$ can be the same as the training set $S_{\Omega}^k$ for the Deep Ritz method. 
$\mu^{k+1}$ in equation \eqref{A2} can be omitted since it is a constant. After $\theta_f^{k+1}$ is computed, we obtain the bounded KRnet at the $(k+1)$th adaptivity iteration and then use it to sample new training sets $S_{\Omega}^{k+1}$ and $S_{B}^{k+1}$ for the next iteration. 
The whole algorithm is shown in Algorithm $\ref{Adaptive sampling method for Deep Ritz}$.

\begin{algorithm}[H]
\caption{Adaptive sampling method for Deep Ritz}\label{Adaptive sampling method for Deep Ritz}
\begin{algorithmic}[1]
\REQUIRE  Initial $u(x,\theta^{0})$, $p_{\text{bKRnet}}(x,\theta_{f}^{0})$, maximum epoch number $N_{e}$, batch size $m$, initial Deep Ritz training set $S_{\Omega}^{0}$ and $S_{\partial \Omega}^{0}$, initial bounded KRnet training set $S_{B}^{0}$
\FOR{$k=0$ \TO $N_{\text{adaptive}}-1$}
\STATE // Solve PDE by Deep Ritz method
\FOR{$i=0$ \TO $N_{e}$}
\FOR{$j=0$ \TO $J$}
\STATE Sample $m$ samples from $S_{\Omega}^{k}$.
\STATE Sample $m$ samples from $S_{\partial\Omega}^{k}$.
\STATE Update $u(x,\theta)$ by descending the stochastic gradient of the objective in $(\ref{A1})$.
\ENDFOR
\ENDFOR
\STATE // Train bounded KRnet
\FOR{$i=0$ \TO $N_{e}$}
\FOR{$j=0$ \TO $J$}
\STATE Sample $m$ samples from $S_{B}^{k}$
\STATE Update $p_{\text{bKRnet}}(x,\theta_f)$ by descending the stochastic gradient of the objective in $(\ref{A2})$.
\ENDFOR
\ENDFOR
\STATE // Generate new training set
\STATE Generate $S_{\Omega}^{k+1}$ and $S_{B}^{k+1}$ using $p_{\text{bKRnet}}(x,\theta_f^{k+1})$.
\ENDFOR
\end{algorithmic}
\end{algorithm}

\subsubsection{Alternative PDF models and sample generation}
In practice, using the numerical approximation of the optimal PDF for the change of measure may cause a problem that the ratio $\frac{G(u(x,\theta))}{p(x)}$ in equation \eqref{importance sampling} is too large for some samples. This is possible since the numerical approximation of $|G(u(x))|/\mu$ introduces uncertainties where we do not have any control over the ratio. We then consider two alternative PDF models to enhance both the robustness and effectiveness:
\begin{enumerate}
\item
We build a mixture model by combining the PDF induced by bounded KRnet and the PDF of the uniform distribution on the domain. 
\item
We build a mixture model by combining the PDF induced by bounded KRnet in the current adaptivity iteration and the mixture model in the previous adaptivity iteration.
\end{enumerate} 

We look at the first alternative PDF model, which is a mixture model defined as
\begin{equation}\label{alternative_model_1}
p_{\text{mixture}}(x,\theta_f^{k+1})=\epsilon\,p_{\text{bKRnet}}(x,\theta_{f}^{k+1})+(1-\epsilon)\,p_{\text{uniform}}(x),\quad \forall x\in S_{\Omega}^{k+1},
\end{equation}
where $0<\epsilon< 1$. In contrast to the basic algorithm, a portion of the training points are from the uniform distribution instead of bounded KRnet. Noting that
\[
p_{\text{mixture}}(x,\theta_f^{k+1})\geq\frac{1-\epsilon}{|\Omega|},
\]
we then avoid the problem that the ratio $\frac{G(u(x,\theta_f^{k+1}))}{p_{\text{mixture}}(x,\theta_f^{k+1})}$ becomes too large. The same approach can also be applied to the training set $S_{B}^k$. The algorithm is presented in Algorithm $\ref{model 1}$.


\begin{algorithm}[H]
\caption{Adaptive sampling with alternative PDF model $1$}\label{model 1}
\begin{algorithmic}[1]
\REQUIRE  Initial $u(x,\theta^{0})$, $p_{\text{bKRnet}}(x,\theta_{f}^{0})$, maximum epoch number $N_{e}$, batch size $m$, initial Deep Ritz training set $S_{\Omega}^{0}$ and $S_{\partial \Omega}^{0}$, initial bounded KRnet training set $S_{B}^{0}$, mixture parameter $\epsilon$
\FOR{$k=0$ \TO $N_{\text{adaptive}}-1$}
\STATE // Solve PDE by Deep Ritz method
\FOR{$i=0$ \TO $N_{e}$}
\FOR{$j=0$ \TO $J$}
\STATE Sample $m$ samples from $S_{\Omega}^{k}$.
\STATE Sample $m$ samples from $S_{\partial\Omega}^{k}$.
\STATE Update $u(x,\theta)$ by descending the stochastic gradient of the objective in $(\ref{A1})$.
\ENDFOR
\ENDFOR
\STATE // Train bounded KRnet
\FOR{$i=0$ \TO $N_{e}$}
\FOR{$j=0$ \TO $J$}
\STATE Sample $m$ samples from $S_{B}^{k}$
\STATE Update $p_{\text{bKRnet}}(x,\theta_f)$ by descending the stochastic gradient of the objective in $(\ref{A2})$.
\ENDFOR
\ENDFOR
\STATE // Generate new training set
\STATE Generate uniform training sets $S_{\Omega,\text{uniform}}^{k+1}$ and $S_{B,\text{uniform}}^{k+1}$ using the uniform distribution on $\Omega$.
\STATE Generate training sets  $S_{\Omega,\text{bKRnet}}^{k+1}$ and $S_{B,\text{bKRnet}}^{k+1}$ using $p_{\text{bKRnet}}(x,\theta_f^{k+1})$.
\STATE Set $S_{\Omega}^{k+1}=S_{\Omega,\text{uniform}}^{k+1}\cup S_{\Omega,\text{bKRnet}}^{k+1}$ and $S_{B}^{k+1}=S_{B,\text{uniform}}^{k+1}\cup S_{B,\text{bKRnet}}^{k+1}$.
\STATE Calculate the PDF values of training sets $S_{\Omega}^{k+1}$ and $S_{B}^{k+1}$ using $(\ref{alternative_model_1})$.
\ENDFOR
\end{algorithmic}
\end{algorithm}

We now present the second alternative PDF model, 
which is defined as
\begin{equation}\label{method_2}
\begin{aligned}
&p_{\text{mixture}}(x,\left\{\theta_f^{t}, t\le k+1\right\})\\
=&\epsilon\,p_{\text{bKRnet}}(x,\theta_{f}^{k+1})+(1-\epsilon)\,p_{\text{mixture}}(x,\left\{\theta_{f}^{t},t\le k\right\})\\
=&\epsilon\,p_{\text{bKRnet}}(x,\theta_{f}^{k+1})\\
&+(1-\epsilon)\left(\epsilon\,p_{\text{bKRnet}}(x,\theta_{f}^{k})+(1-\epsilon)\,p_{\text{mixture}}(x,\left\{\theta_{f}^{t},t\le k-1\right\})\right)\\
=&\cdots\\
=&\sum_{t=1}^{k+1}\epsilon (1-\epsilon)^{k+1-t}\,p_{\text{bKRnet}}(x,\theta_{f}^{t})+(1-\epsilon)^{k+1}/|\Omega|
\end{aligned}
\end{equation}
where we form a mixture model using the PDF model $p_{\text{mixture}}(x,\left\{\theta_{f}^{t},t\le k\right\})$ at adaptivity iteration $k$ and the PDF model $p_{\text{bKRnet}}(x,\theta_{f}^{k+1})$ at adaptivity iteration $k+1$. The initial data set $S_{\Omega}^0$ is sampled from a uniform distribution. As shown in  equation \eqref{method_2}, the training points from the last mixture model and their PDF values depend on all the previous bounded KRnets. 
The value of the proposed mixture model is at least $(1-\epsilon)^{k+1}/|\Omega|$. 
The same manipulation can also be implemented to the training set $S_{B}^{k+1}$. 

The second alternative PDF model is an extension of the first one. In the second alternative PDF model, all previous bounded KRnets are saved and reused every iteration. 
Although method $2$ needs more storage and computation cost, it is more flexible than method $1$ and the extra cost can be neglected compared to the total cost. The algorithm is presented in Algorithm $\ref{model 2}$.

\begin{algorithm}[H]
\caption{Adaptive sampling with alternative PDF model $2$}\label{model 2}
\begin{algorithmic}[1]
\REQUIRE  Initial $u(x,\theta^{0})$, $p_{\text{bKRnet}}(x,\theta_{f}^{0})$, maximum epoch number $N_{e}$, batch size $m$, initial Deep Ritz training set $S_{\Omega}^{0}$ and $S_{\partial \Omega}^{0}$, initial bounded KRnet training set $S_{B}^{0}$, mixture parameter $\epsilon$
\FOR{$k=0$ \TO $N_{\text{adaptive}}-1$}
\STATE // Solve PDE by Deep Ritz method
\FOR{$i=0$ \TO $N_{e}$}
\FOR{$j=0$ \TO $J$}
\STATE Sample $m$ samples from $S_{\Omega}^{k}$.
\STATE Sample $m$ samples from $S_{\partial\Omega}^{k}$.
\STATE Update $u(x,\theta)$ by descending the stochastic gradient of the objective in $(\ref{A1})$.
\ENDFOR
\ENDFOR
\STATE // Train bounded KRnet
\FOR{$i=0$ \TO $N_{e}$}
\FOR{$j=0$ \TO $J$}
\STATE Sample $m$ samples from $S_{B}^{k}$
\STATE Update $p_{\text{bKRnet}}(x,\theta_f)$ by descending the stochastic gradient of the objective in $(\ref{A2})$.
\ENDFOR
\ENDFOR
\STATE // Save bounded KRnet 
\STATE Restore current bound KRnet parameters $\theta_f^{k+1}$.
\STATE // Generate new training set
\STATE Initialize new training sets with uniform training sets $S_{\Omega}^{k+1}=S_{\Omega,\text{uniform}}^{k+1}$ and $S_{B}^{k+1}=S_{B,\text{uniform}}^{k+1}$.
\FOR{$t=0$ \TO $k+1$}
\STATE Load bounded KRnet with parameters $\theta_f^{t+1}$.
\STATE Generate training sets $S_{\Omega,\text{bKRnet}}^{t+1}$, $S_{B,\text{bKRnet}}^{t+1}$ using $p_{\text{bKRnet}}(x,\theta_f^{t+1})$.
\STATE Set $S_{\Omega}^{k+1}=S_{\Omega}^{k+1}\cup S_{\Omega,\text{bKRnet}}^{t+1}$ and $S_{B}^{k+1}=S_{B}^{k+1}\cup S_{B,\text{bKRnet}}^{t+1}$  .
\ENDFOR
\STATE Calculate the PDF values of training sets $S_{\Omega}^{k+1}$ and $S_{B}^{k+1}$ using $(\ref{method_2})$.
\ENDFOR
\end{algorithmic}
\end{algorithm}

\section{Numerical experiments}\label{Numerical experiment}
In this section, we present some numerical experiments to illustrate the effectiveness of our methods. We test four examples; among these examples, two have a solution with low regularity, another has a singularity in the solution, and the last one is a high-dimensional problem. For the Deep Ritz method, we use a deep network structure similar to ResNet~\cite{he2016deep}. 
In~\cite{chen2020comparison}, the authors compared the performance of several different activation functions. Their results show that $\sin^3(x)$ is the best among all tested activation functions. Hence, we use $\sin^3(x)$ as the activation function in all experiments. For bounded KRnet, we use the same architecture as described in~\cite{zeng2023bounded}. 
For both Deep Ritz method and bounded KRnet, we train the neural networks by the ADAM method. In order to test the validity of our method, we use the following $L_2$ error 
$$
\text{Error}=\frac{\sqrt{\sum_{i=1}^N|u(x_i,\theta)-u(x_i)|^2}}{\sqrt{\sum_{i=1}^N|u(x_i)|^2}}
$$
in our experiments where the set of test points is either generated using a tensor grid or drawn from a uniform distribution (for high dimensional problem).  In the original Deep Ritz method, the training points are reselected from a uniform distribution at each training epoch. Hence, the concept total number of training points is not explicitly defined for the Deep Ritz method since it depends on the training process. One typically uses the batch size (the number of training points used in one training epoch) to describe the scale of the training set in the Deep Ritz method. Similar to the original Deep Ritz method, in our adaptive approach, new training points can be efficiently sampled from a bounded KRnet at each training epoch for the basic algorithm and the first alternative PDF model. For the second alternative PDF model, the dynamic sampling is not efficient since all previous models are involved. In this work, at the $k$th adaptive iteration, we generate a finite training set $S_{\Omega}^k$($S_{B}^k$) of size $N_v$($N_B$), based on which mini-batches are formed for each training epoch. 
Across all numerical experiments, we compare the performance of the following four sampling strategies:
(1) The conventional Deep Ritz, where the training set will be completely updated with new uniform samples at each adaptivity iteration;
(2) The basic adaptive algorithm based on bounded KRnet, i.e., Algorithm $\ref{Adaptive sampling method for Deep Ritz}$;
(3) The adaptive algorithm based on the first alternative PDF model, i.e., Algorithm $\ref{model 1}$; 
and (4) The adaptive algorithm based on the second alternative PDF model, i.e., Algorithm $\ref{model 2}$. 
For a fair comparison, we use the same uniformly distributed training set, denoted as $S_{\Omega}^{0}$ for Deep Ritz at iteration $0$ across all four sampling strategies. Similarly, we employ the same uniformly distributed training set, denoted as $S_{B}^0$, to train bounded KRnet at iteration $0$ for all three bKRnet-based methods.

\subsection{Two dimensional peak problem}
We consider the following 2D elliptic problem with a single peak:
$$
\begin{aligned}
-\Delta u(x_1,x_2)&=f(x_1,x_2), &(x_1,x_2)&\in \Omega\\
u(x_1,x_2)&=g(x_1,x_2), &(x_1,x_2)&\in \partial \Omega
\end{aligned}
$$
where the computational domain is $ \Omega=[-1,1]^2$.
The exact solution is given by 
$$
u(x_1,x_2)=\exp(-1000((x_1-0.5)^2+(x_2-0.5)^2))
$$
which has a peak at $(0.5, 0.5)$ and decreases rapidly away from $(0.5, 0.5)$.
This is a classical benchmark problem commonly used to assess the performance of adaptive techniques in both deep learning methods~\cite{tang2023pinns, gao2023failure} and finite element methods~\cite{morin2002convergence}.
\newpage
\begin{figure}[H]
\begin{center}
\subfigure[Calculated solution, True solution, Absolute error]{
\scalebox{0.35}[0.35]{\includegraphics{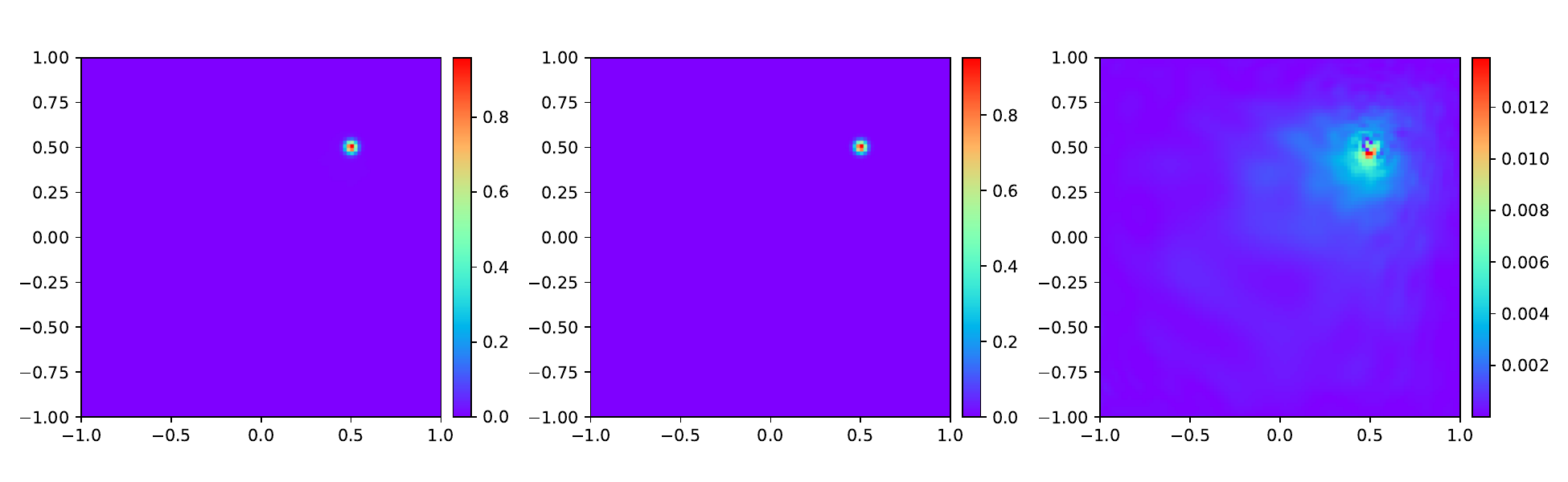}}
}
\end{center}
\caption{Numerical results of the alternative model $1$ with $\epsilon=0.5$: Numerical solution, Exact solution, Absolute error  at iteration $7$.}\label{numerical_result_one_peak}
\end{figure}

We choose a ResNet $u(x,\theta)$ with a stack of $4$ blocks(eight fully-connected layers) and $32$ neurons in each layers to approximate the solution. For bounded KRnet, we use $6$ CDF layers and two fully connected layers with $24$ neurons for each CDF layers. The training process for the Deep Ritz method consists of $10000$ epochs for the initial iteration, followed by an additional $1000$ epochs for each subsequent iteration. The number of epochs for training the bounded KRnet is $3000$. The balance parameter $\beta$ in the Deep Ritz method is configured to be $100$. The learning rate for ADAM operator is set to $0.0001$ and the batch size is set to $m=5000$. We test our problem on a $101\times 101$ tensor grid.

\begin{figure}[H]
\begin{center}
\subfigure[Training points at iteration $1$]{
\scalebox{0.25}[0.3125]{\includegraphics{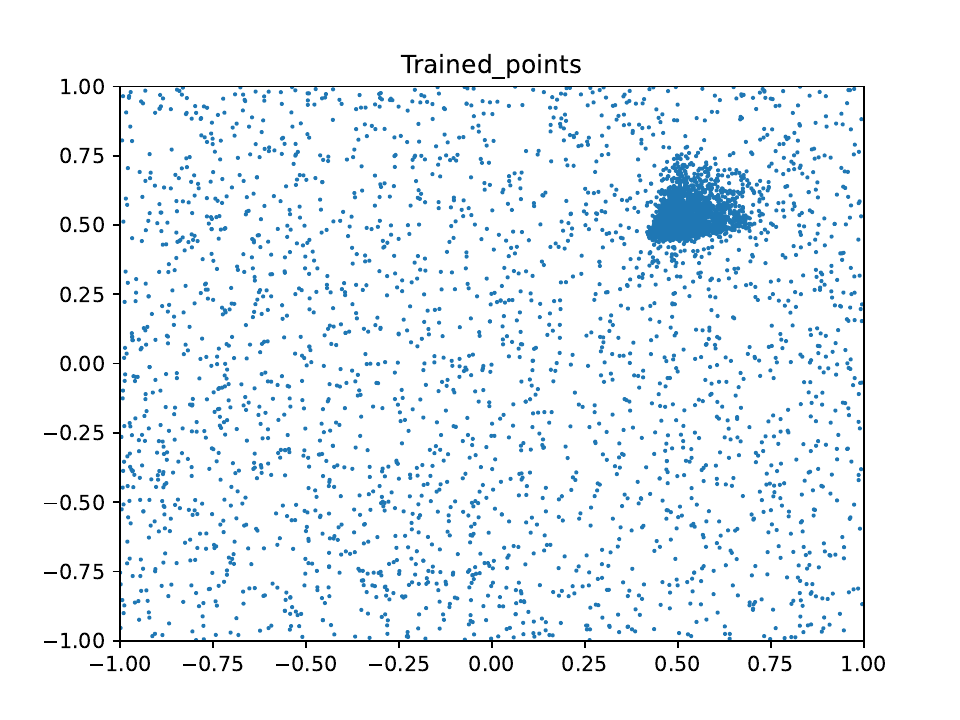}}
}
\hspace{10mm}
\subfigure[Training points at iteration $3$]{
\scalebox{0.25}[0.3125]{\includegraphics{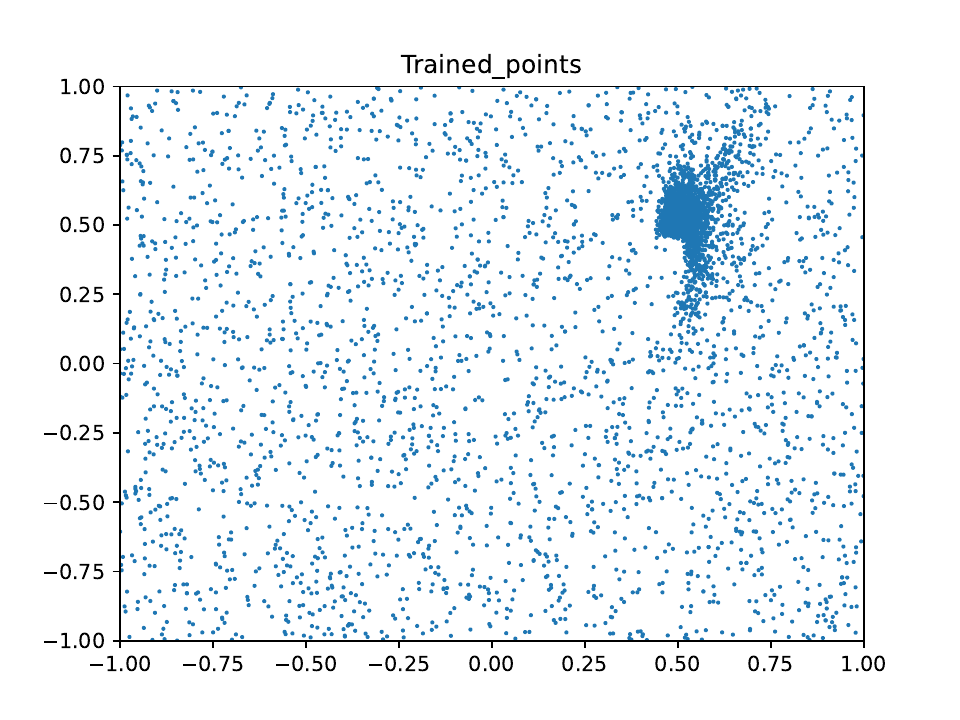}}
}

\subfigure[Training points at iteration $5$]{
\scalebox{0.25}[0.3125]{\includegraphics{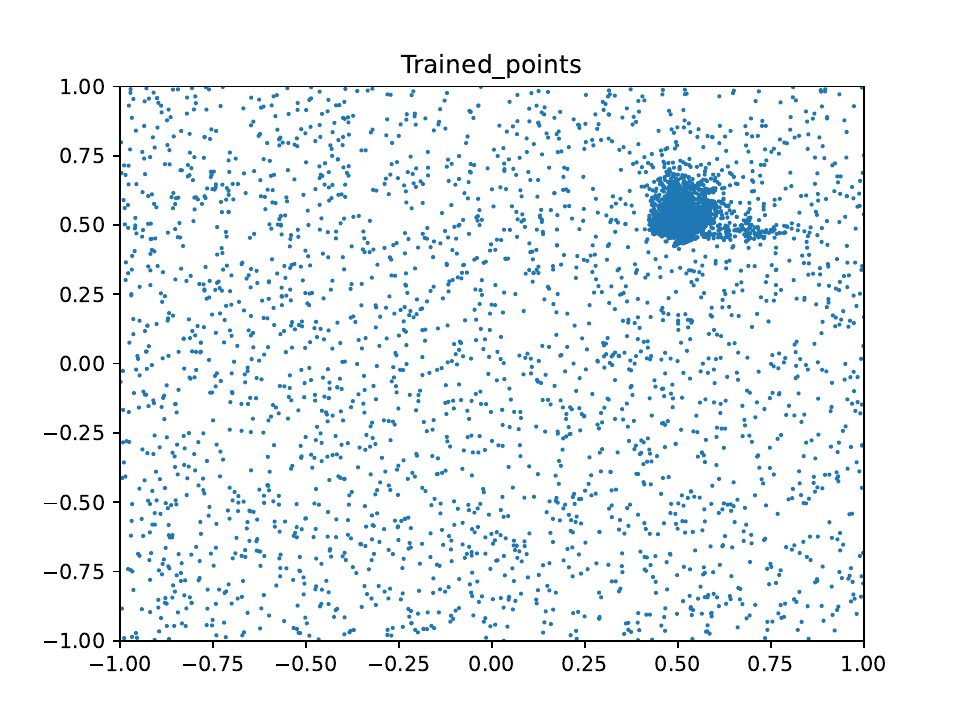}}
}
\hspace{10mm}
\subfigure[Training points at iteration $7$]{
\scalebox{0.25}[0.3125]{\includegraphics{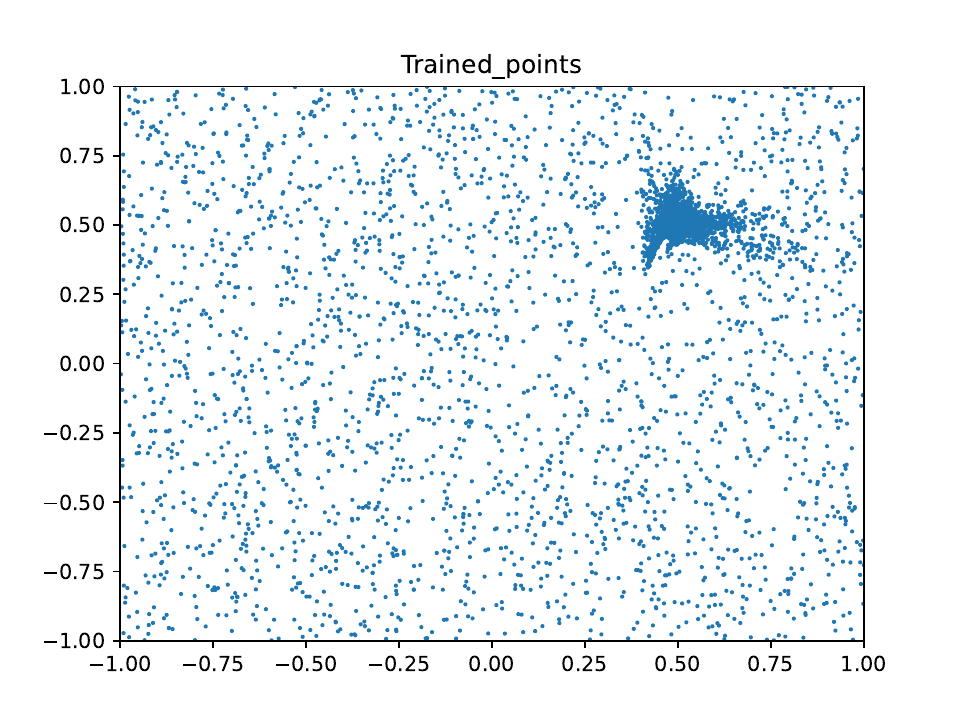}}
}
\end{center}
\caption{Training points of the alternative model $1$ with $\epsilon=0.5$}\label{train_points_one_peak}
\end{figure}

We first apply the alternative PDF model $1$ with $\epsilon=0.5$ to the Deep Ritz method. The numerical results are presented in Figure $\ref{numerical_result_one_peak}$.  We show the evolution of the training sets with respect to adaptivity iterations $k=1,3,5,7$ in Figure $\ref{train_points_one_peak}$. It can be observed that the peak density in each training set consistently occurs around the coordinates $[0.5, 0.5]$.
This is because we use bKRnet to approximate the normalized absolute integrand(computed from the Deep Ritz solution), and its shape remains relatively stable during the training process. Note that this is different from the distributions of training points generated by the residual based DAS-R for PINNs. The distribution of training points generated by DAS-R becomes less concentrated as the number of adaptive iteration increases. Another difference is that computing the residual during bKRnet training in DAS-R requires second order derivatives while only first order derivative are needed in our approach. 

\begin{figure}[H]
\begin{center}
\subfigure[Alternative model $1$ with $\epsilon=0.5$]{
\scalebox{0.35}[0.35]{\includegraphics{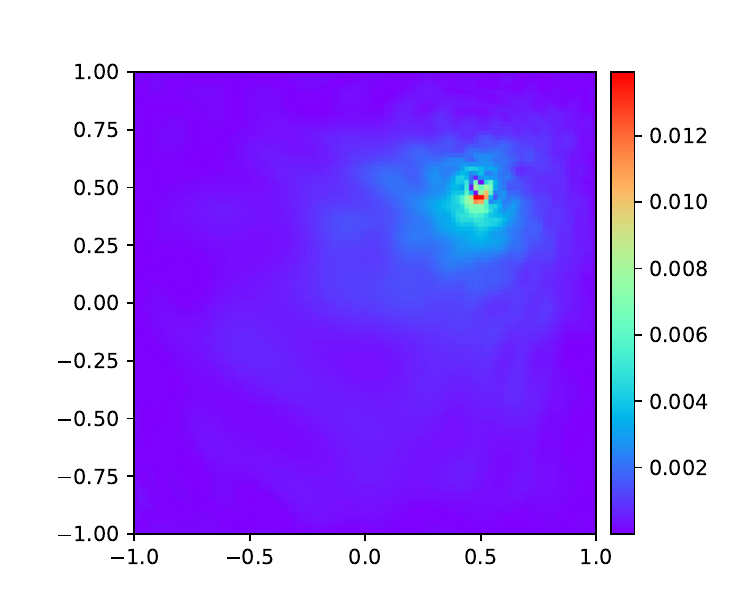}}
}
\hspace{10mm}
\subfigure[Alternative model $2$ with $\epsilon=0.3$]{
\scalebox{0.35}[0.35]{\includegraphics{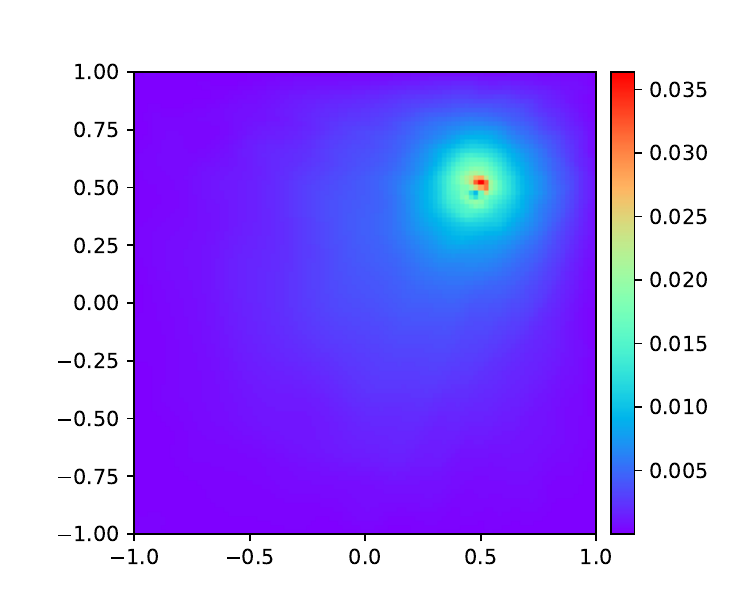}}
}

\subfigure[Uniform]{
\scalebox{0.35}[0.35]{\includegraphics{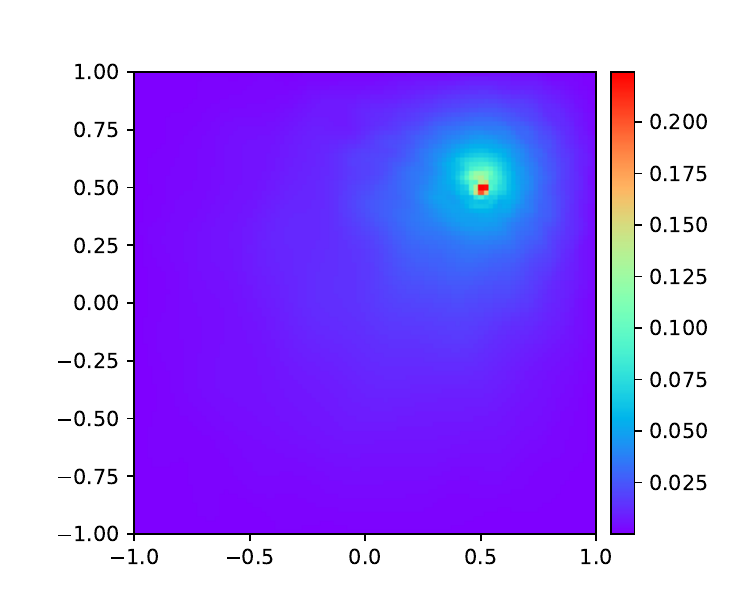}}
}
\hspace{10mm}
\subfigure[Pure bounded KRnet]{
\scalebox{0.35}[0.35]{\includegraphics{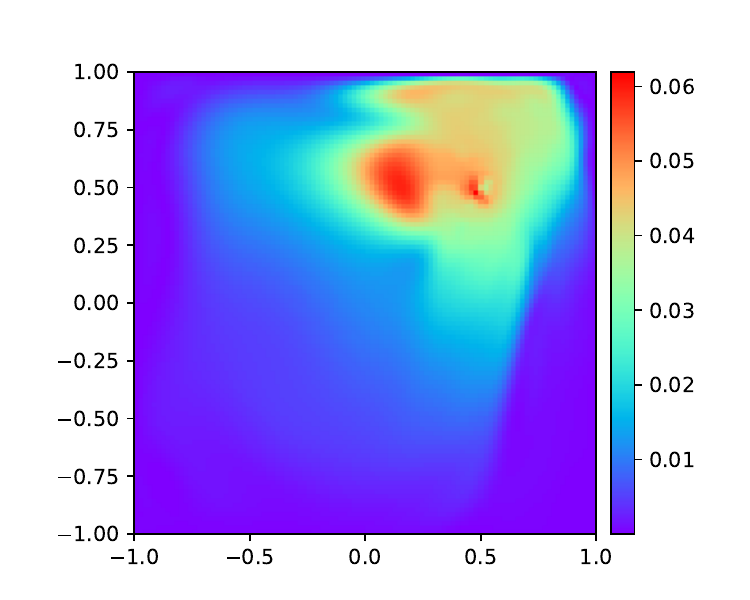}}
}
\end{center}
\caption{Comparison of numerical solutions using different sampling schemes}\label{numerical_solution_comp_one_peak}
\end{figure}

We compare the performances of different adaptive sampling strategies in Figure $\ref{numerical_solution_comp_one_peak}$.  The size of the training set is 200000. It is evident that all strategies based on importance sampling can capture information in the region of low regularity more effectively than the original Deep Ritz method. 

\begin{figure}[H]
\begin{center}
\subfigure[Relative $L_2$ error]{
\scalebox{0.5}[0.5]{\includegraphics{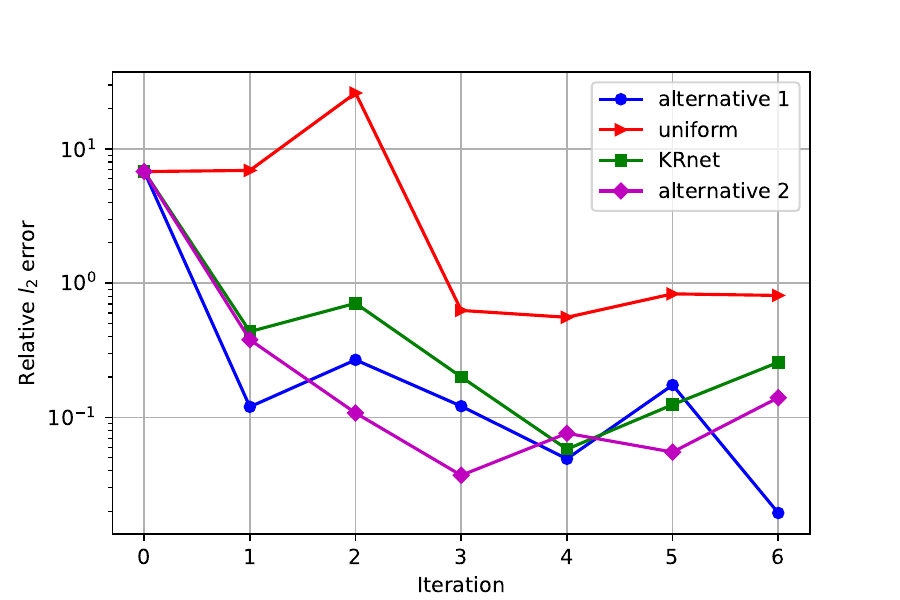}}
}
\end{center}
\caption{Relative $L_2$ errors}\label{Error_one_peak}
\end{figure}

In Figure $\ref{Error_one_peak}$, we present the $L_2$ errors for four sampling strategies. As all adaptive sampling methods require replacing all training points at each iteration, the relative $L_2$ errors do not exhibit a monotonic decrease. However, a decreasing trend can still be observed, allowing us to compare the performances of different methods. Among all the methods, the alternative model $1$  provides the most accurate solution. It is also worth noting that the alternative method $2$ provides a reasonably accurate solution at iteration $3$ which means we can stop our adaptive method earlier to save the computational cost. 

\subsection{Two dimensional problem with two peaks}
We consider the following two dimensional problem with two peaks
$$
\begin{aligned}
-\nabla^2u(x_1,x_2)+\nabla\cdot [u(x_1,x_2)\nabla(x_1^2+x_2^2)]&=s(x_1,x_2), &(x_1,x_2) &\in \Omega\\
u(x_1,x_2)&=g(x_1,x_2),&(x_1,x_2) &\in \partial \Omega.
\end{aligned}
$$
where the computation domain is $\Omega=[-1,1]^2$. We choose the exact solution of the problem to be
$$
\begin{aligned}
u(x_1,x_2)=&\exp(-1000[(x_1-0.5)^2+(x_2-0.5)^2])\\
&+\exp(-1000[(x_1+0.5)^2+(x_2+0.5)^2])
\end{aligned}
$$
which has two peaks at the points $(0.5,0.5)$ and $(-0.5,-0.5)$. The Dirichlet boundary condition $g(x_1,x_2)$ and the right-hand side term are also given by the exact solution.

\begin{figure}[H]
\begin{center}
\subfigure[Calculated solution, True solution, Absolute error]{
\scalebox{0.35}[0.35]{\includegraphics{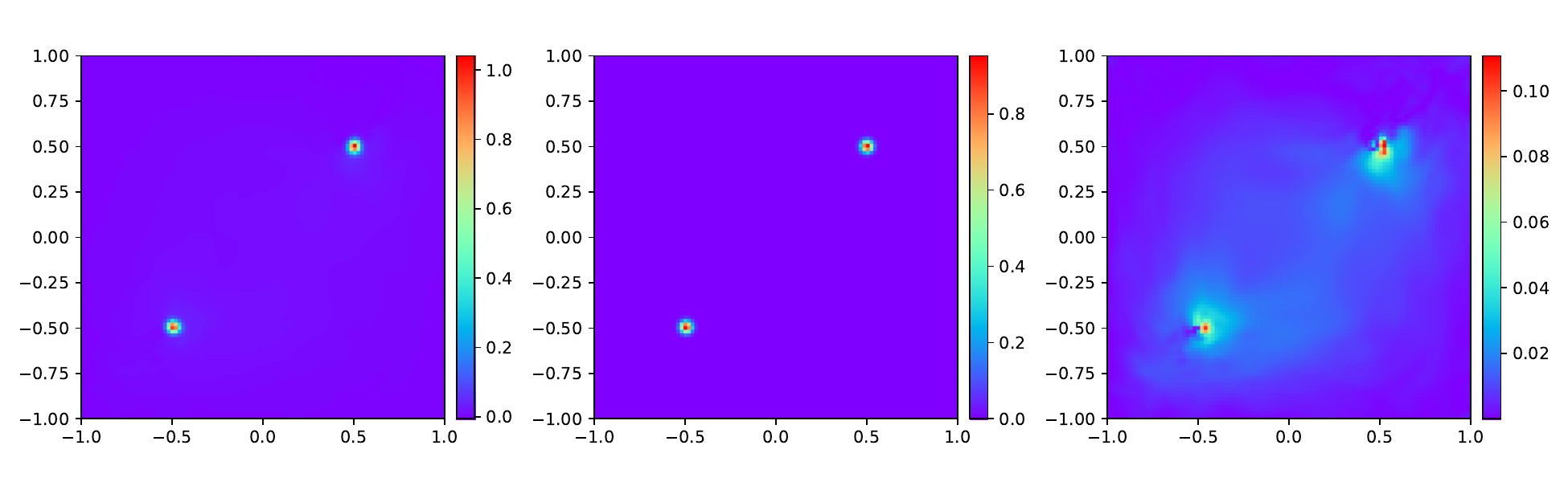}}
}
\end{center}
\caption{Numerical results of the alternative model $2$ with $\epsilon=0.1$: Numerical solution, Exact solution, Absolute error  at iteration $7$}\label{numerical_result_two_peak}
\end{figure}

We apply the same ResNet and bounded KRnet as in the one peak problem. We also utilize the same optimizer and learning rate for training these deep neural networks.  However, in this experiment, we use fewer training epochs and points.

\begin{figure}[H]
\begin{center}
\subfigure[Training points at iteration $1$]{
\scalebox{0.25}[0.3125]{\includegraphics{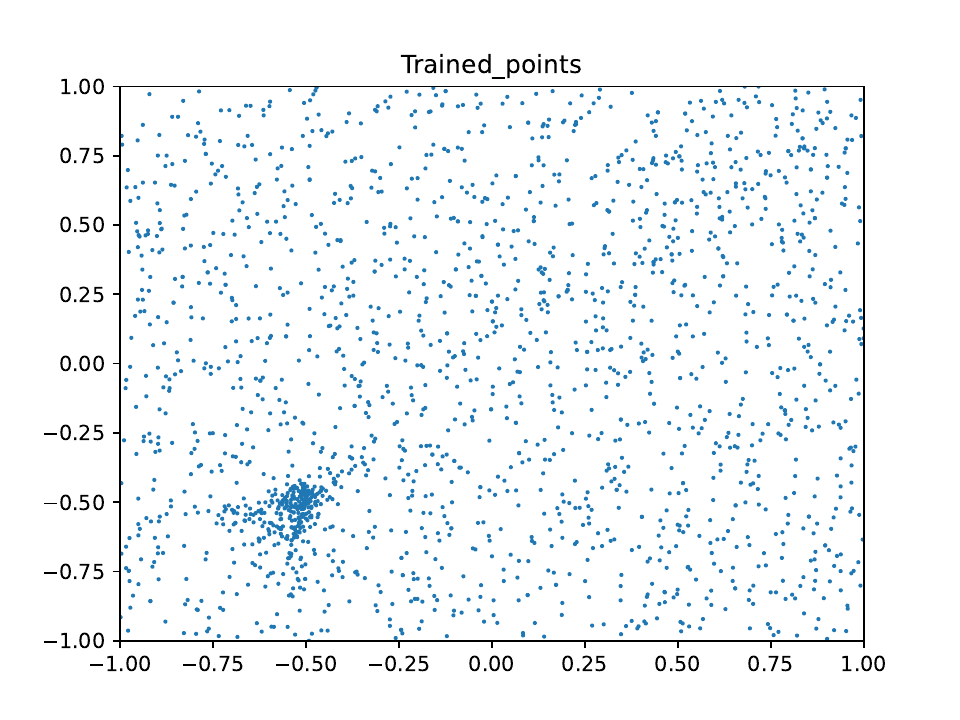}}
}
\hspace{10mm}
\subfigure[Training points at iteration $3$]{
\scalebox{0.25}[0.3125]{\includegraphics{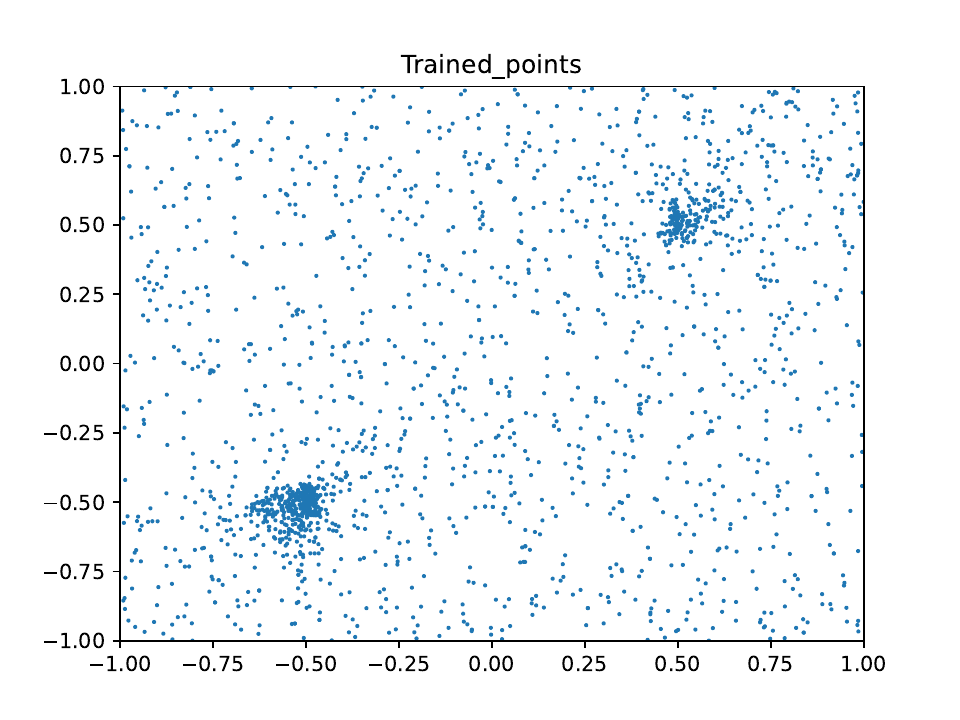}}
}

\subfigure[Training points at iteration $5$]{
\scalebox{0.25}[0.3125]{\includegraphics{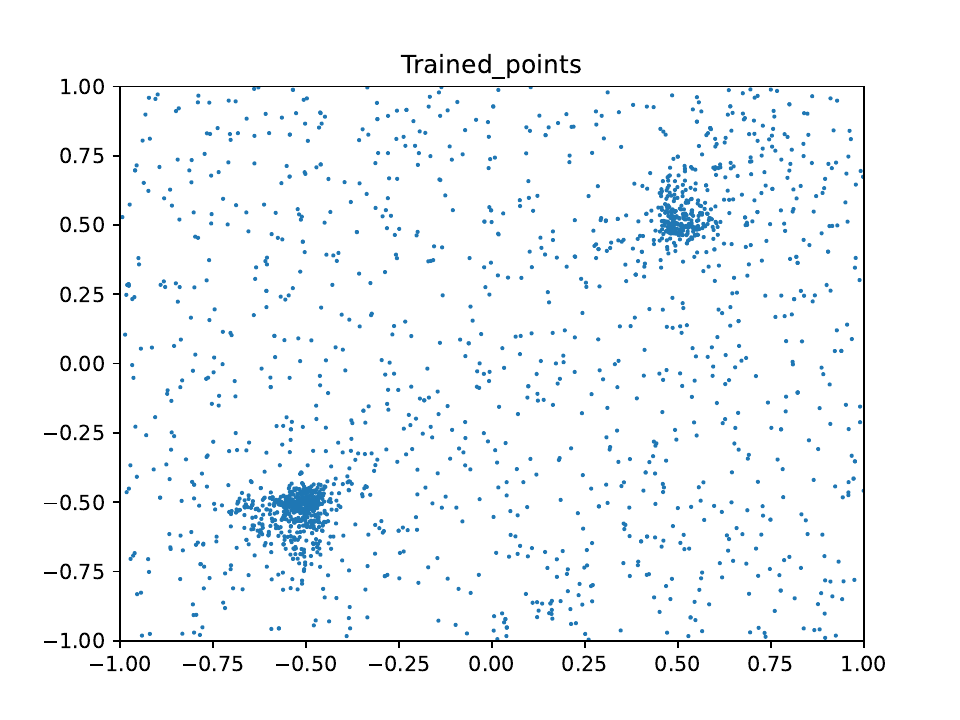}}
}
\hspace{10mm}
\subfigure[Training points at iteration $7$]{
\scalebox{0.25}[0.3125]{\includegraphics{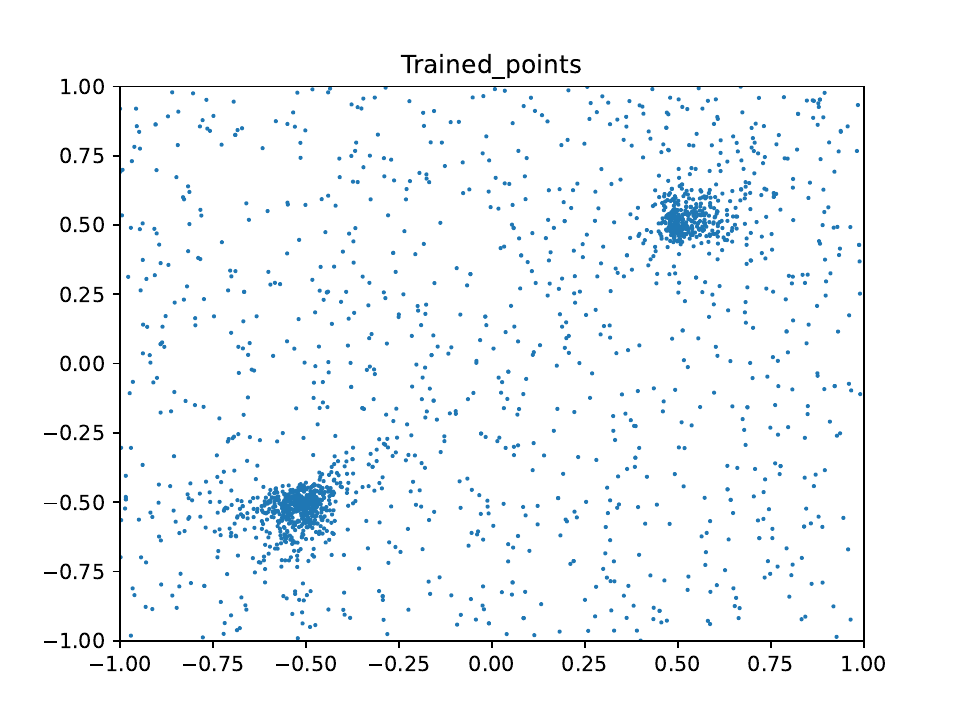}}
}
\end{center}
\caption{Training points of the alternative model $2$ with $\epsilon=0.1$}\label{train_points_two_peak}
\end{figure}

We aim to find out if our adaptive sampling methods can compute a reasonably accurate solution when the uniform sampling strategy fails. In this experiment, the training process for the Deep Ritz method consists of only $3000$ epochs for the initial iteration, followed by an additional $100$ epochs for each subsequent iteration. The number of epochs for training the bounded KRnet is reduced to $1000$.  We have reduced the size of the training set to $20000$ and the batch size to $2000$.

\begin{figure}[H]
\begin{center}
\subfigure[Alternative model $1$ with $\epsilon=0.5$]{
\scalebox{0.35}[0.35]{\includegraphics{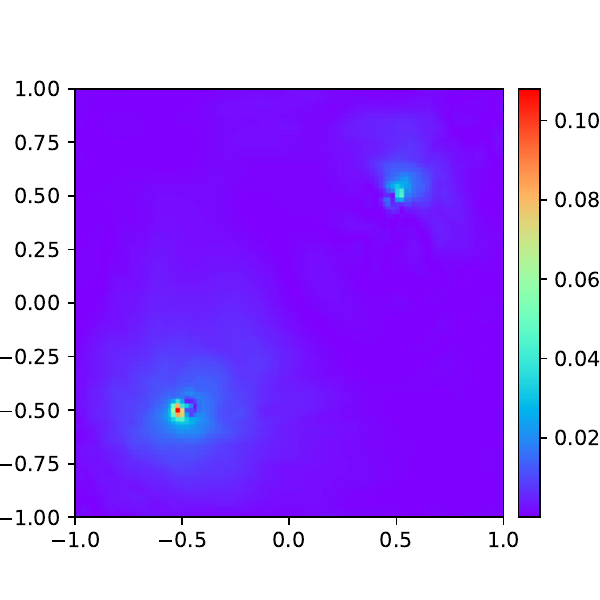}}
}
\hspace{10mm}
\subfigure[Alternative model $2$ with $\epsilon=0.1$]{
\scalebox{0.35}[0.35]{\includegraphics{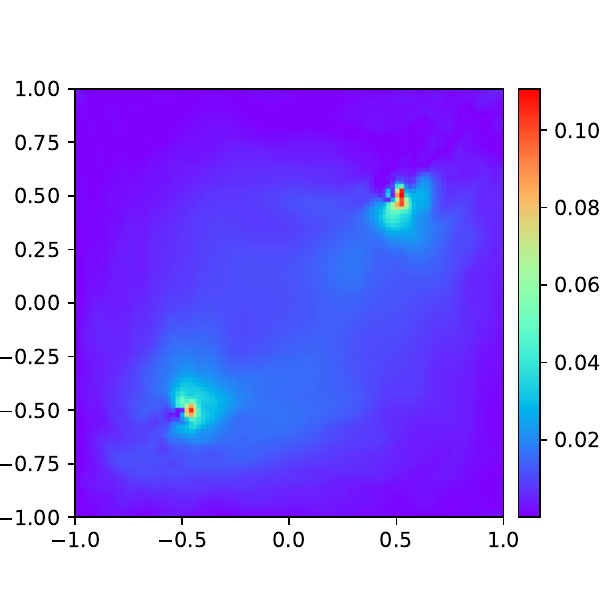}}
}

\subfigure[Uniform]{
\scalebox{0.35}[0.35]{\includegraphics{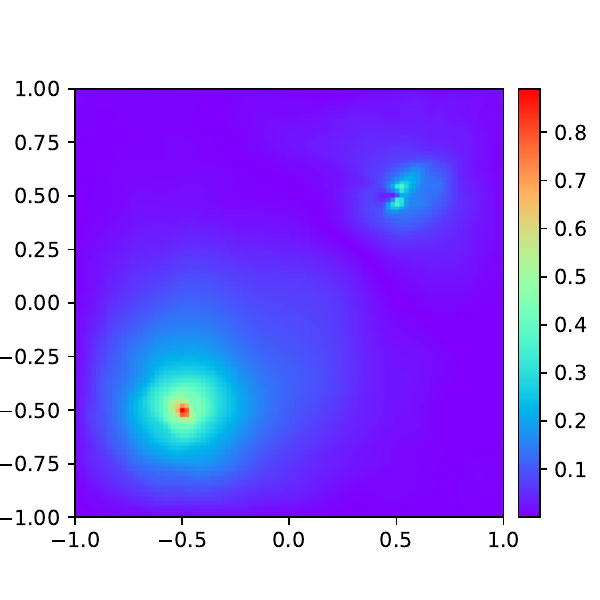}}
}
\hspace{10mm}
\subfigure[Pure bounded KRnet]{
\scalebox{0.35}[0.35]{\includegraphics{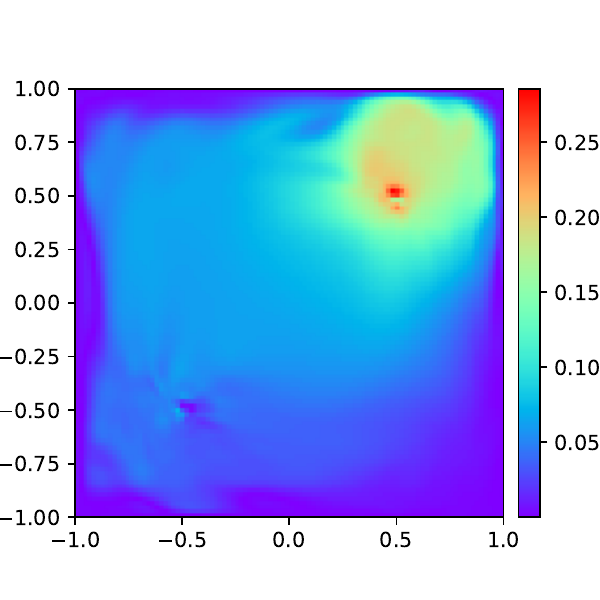}}
}
\end{center}
\caption{Comparison of the absolute errors of numerical solution using different sampling schemes}\label{numerical_solution_comp_two_peak}
\end{figure}

We present the numerical solution and the evolution of training points of the adaptive sampling method based on alternative PDF model $2$ in Figure $\ref{numerical_result_two_peak}$ and Figure $\ref{train_points_two_peak}$ respectively. It is seen that our adaptive methods can effectively capture the information around the two peaks and therefore result in a solution with reasonable accuracy. We compare the performances of four different adaptive sampling strategies in Figure $\ref{numerical_solution_comp_two_peak}$. Due to the complexity of the two peak problem and the reduced number of training epochs and points, the uniform sampling strategy produces an incorrect numerical solution.  In contrast, all three adaptive methods yield reasonably accurate solutions. Among these methods, the two alternative models outperform the pure bounded KRnet.

\subsection{Two-dimensional problem with singularity}
\begin{figure}[H]
\begin{center}
\scalebox{0.35}[0.35]{\includegraphics{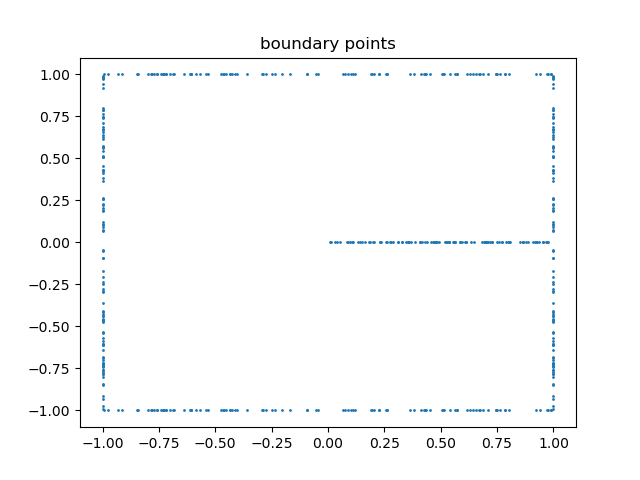}}
\end{center}
\caption{Training points on the boundary of an elliptic problem with singularity.}\label{boundary_singularity}
\end{figure}

We consider the following 2D elliptic problem
$$
\begin{aligned}
-&\Delta u(x_1,x_2)=0, &(x_1,x_2)&\in \Omega\\
&u(x_1,x_2)=u(r,\theta)=r^{\frac{1}{2}}\sin{\frac{\theta}{2}}, &(x_1,x_2)&\in \partial\Omega
\end{aligned}
$$ 
where the domain $\Omega=(-1,1)\times (-1,1)\backslash [0,1)\times \left\{0\right\}$. The boundary is shown in the Figure $\ref{boundary_singularity}$. The solution to this problem suffers from the 'corner singularity' caused by the nature of the domain~\cite{strang1974analysis}. The explicit solution $u^{*}(x)=r^{\frac{1}{2}}\sin\frac{\theta}{2}$ in polar coordinates. 

Considering the problem in the coordinates of $x_1$ and $x_2$, we obtain the following PDE and its associated boundary conditions:
$$
\begin{aligned}
-&\Delta u(x_1,x_2)=0, &(x_1,x_2)&\in \Omega,\\
&u(x_1,x_2)=(x_1^2+x_2^2)^{\frac{1}{4}}\sin\left({\frac{1}{2}\arctan\frac{x_2}{x_1}}\right), &(x_1,x_2)&\in \partial\Omega \cap Q_1,\\
&u(x_1,x_2)=(x_1^2+x_2^2)^{\frac{1}{4}}\sin\left(\pi+{\frac{1}{2}\arctan\frac{x_2}{x_1}}\right), &(x_1,x_2)&\in \partial\Omega \cap Q_2,\\
&u(x_1,x_2)=(x_1^2+x_2^2)^{\frac{1}{4}}\sin\left(\pi+{\frac{1}{2}\arctan\frac{x_2}{x_1}}\right), &(x_1,x_2)&\in \partial\Omega \cap Q_3,\\
&u(x_1,x_2)=(x_1^2+x_2^2)^{\frac{1}{4}}\sin\left(2\pi+{\frac{1}{2}\arctan\frac{x_2}{x_1}}\right), &(x_1,x_2)&\in \partial\Omega \cap Q_4\\
\end{aligned}
$$
where $x_1\neq 0$ and $Q_1, \cdots, Q_4$ represent the four quadrants in 2D space. We then use a ResNet $u(x,\theta)$ to approximate the solution $u(x)$. In the experiment, we choose a ResNet $u(x,\theta)$ with a stack of $4$ blocks(eight fully-connected layers) and $10$ neurons in each layers. For bounded KRnet, we use $6$ CDF layers and two fully connected layers with $24$ neurons for each CDF layers.  We implement $9$ adaptivity iterations for this problem  The number of epochs for training bounded KRnet is set to $3000$ and the number of epochs for training $u(x,\theta)$ is chosen as $20000$ for the first iteration and  followed by an additional $1000$ epochs for each subsequent iteration.
The balance parameter $\beta$ in the Deep Ritz method is configured to be $1000$. The learning rate for ADAM operator is set to $0.001$ and the batch size is set to $m=2000$. In order to avoid testing the singularity at the origin, we test our problem on a $102\times 102$ tensor grid.

\begin{figure}[H]
\begin{center}
\subfigure[Iteration $0$]{
\scalebox{0.25}[0.25]{\includegraphics{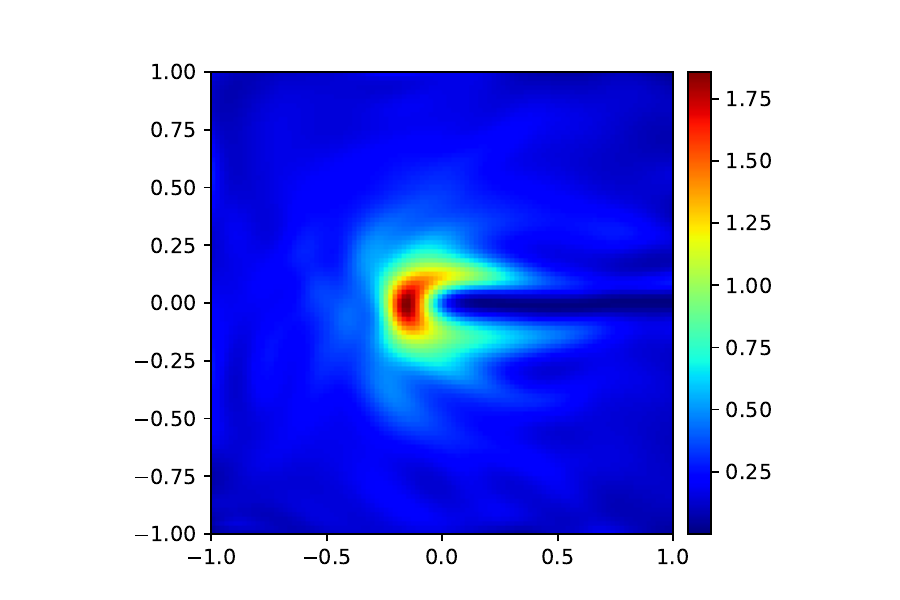}}}
\subfigure[Iteration $2$]{
\scalebox{0.25}[0.25]{\includegraphics{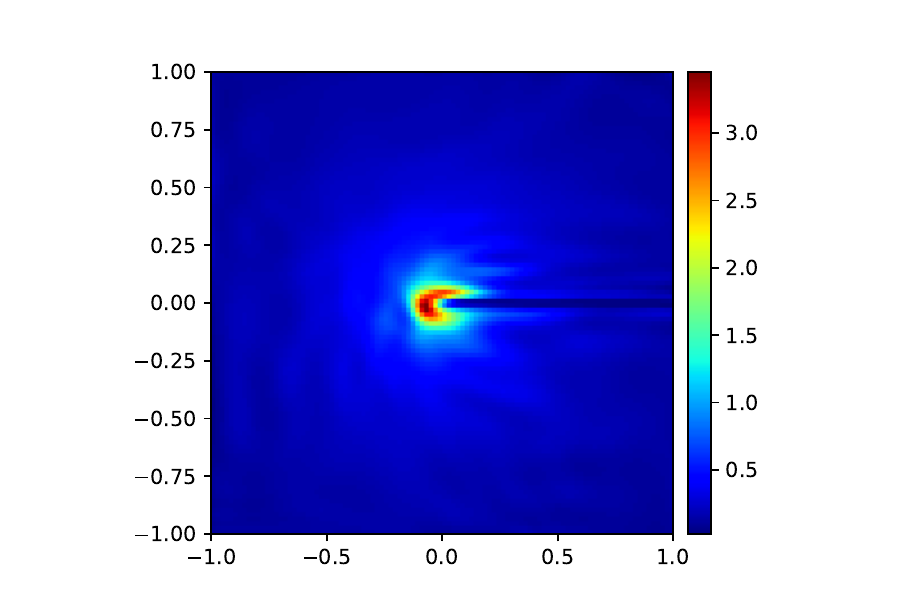}}}
\subfigure[Iteration $4$]{
\scalebox{0.25}[0.25]{\includegraphics{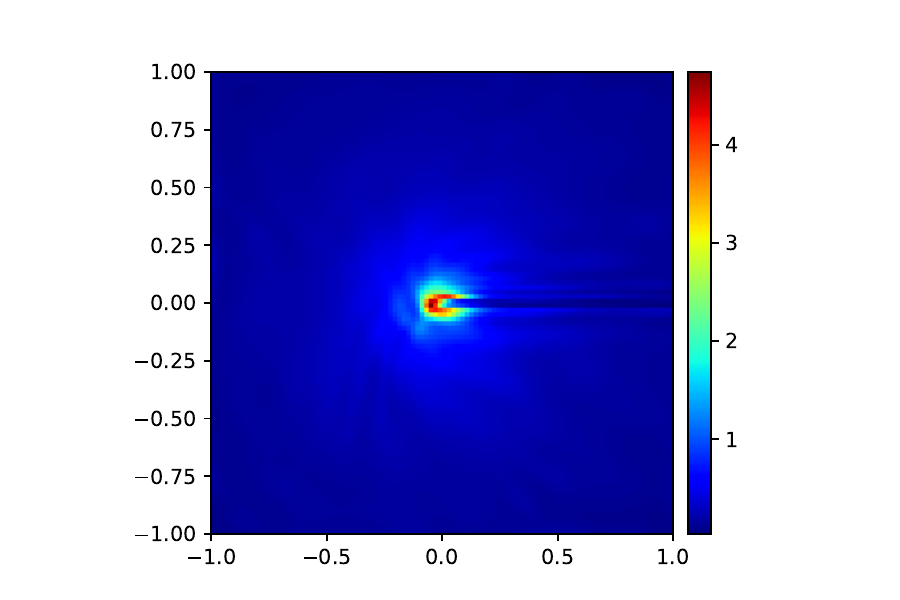}}}

\subfigure[Iteration $6$]{
\scalebox{0.25}[0.25]{\includegraphics{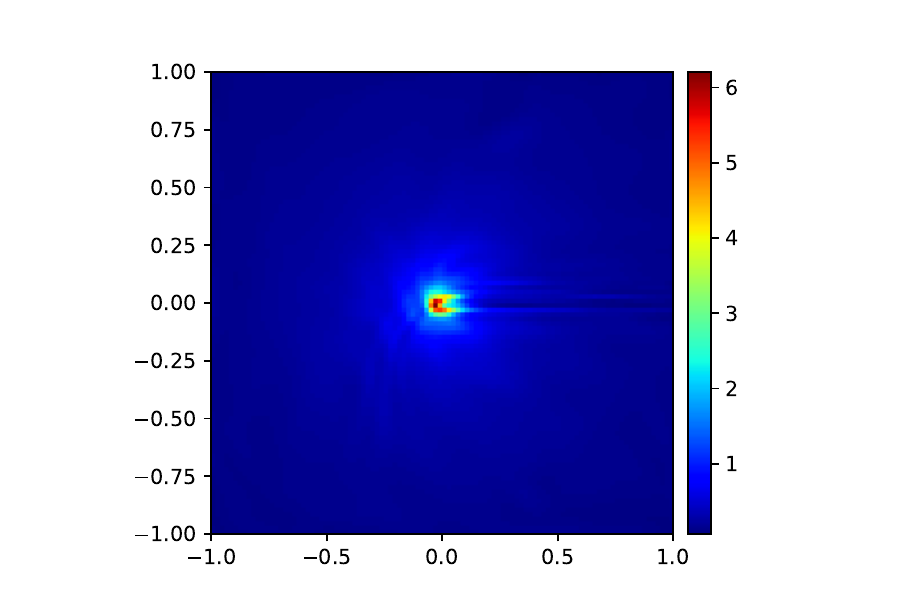}}}
\subfigure[Iteration $8$]{
\scalebox{0.25}[0.25]{\includegraphics{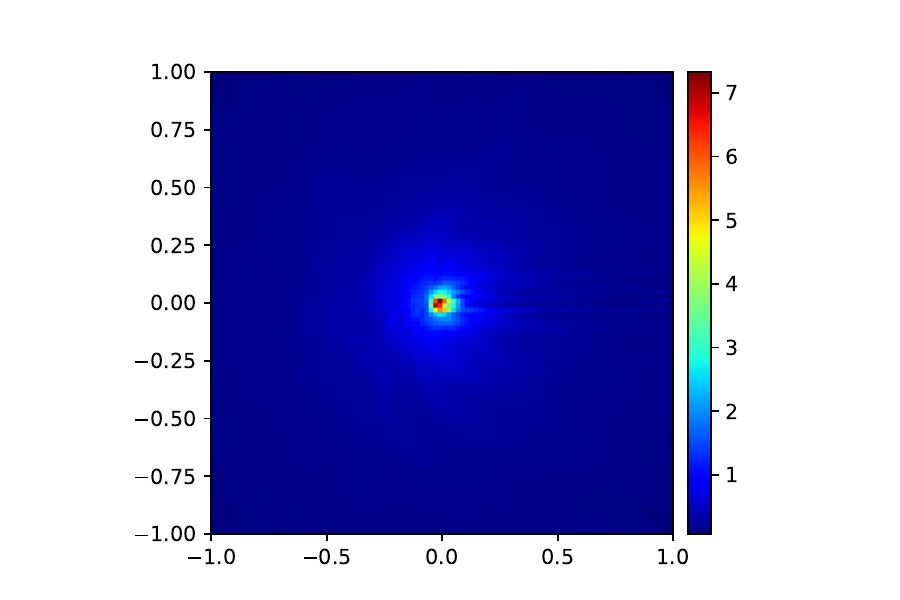}}}
\end{center}
\caption{The evolution of the normalized integrand of the loss functional.}\label{Normalized integrand}
\end{figure}

\begin{figure}[H]
\begin{center}
\subfigure[Iteration $0$]{
\scalebox{0.2}[0.25]{\includegraphics{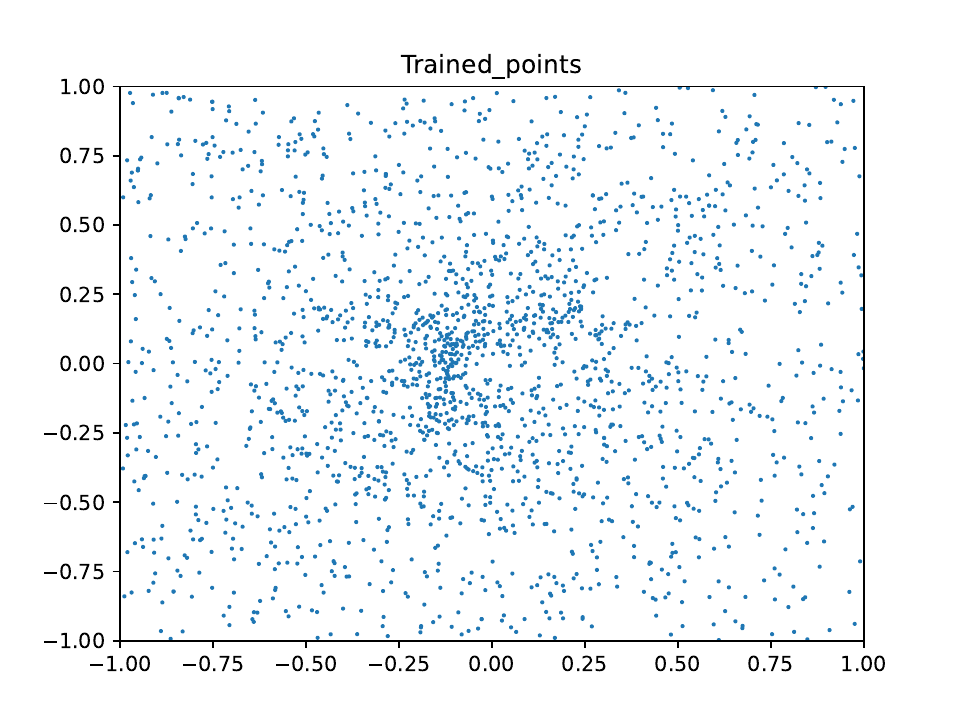}}}
\hspace{5mm}
\subfigure[Iteration $2$]{
\scalebox{0.2}[0.25]{\includegraphics{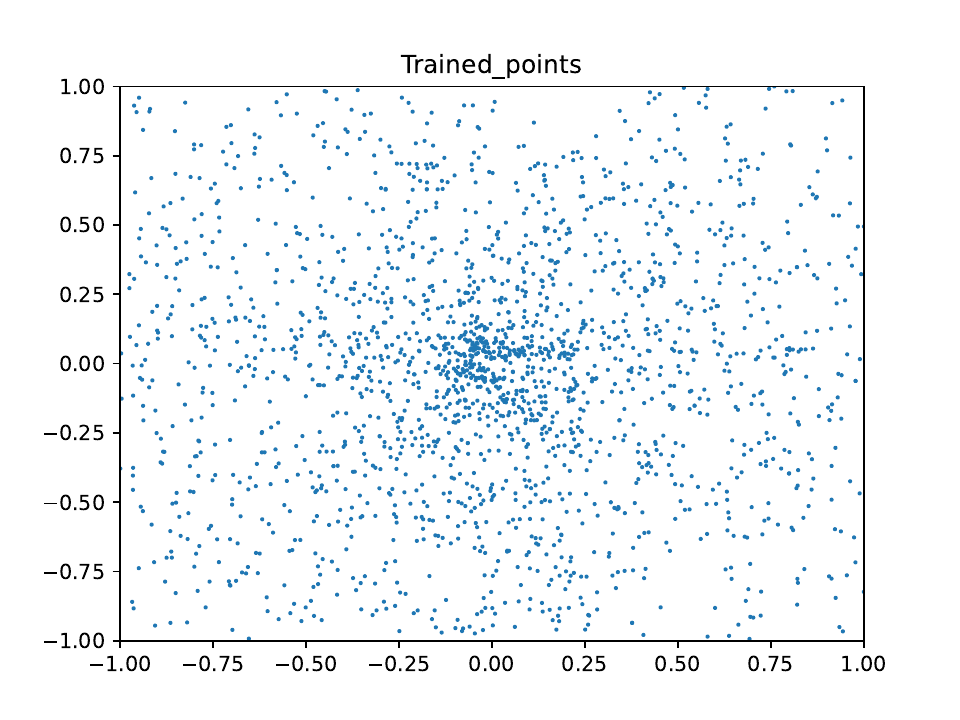}}}
\hspace{5mm}
\subfigure[Iteration $4$]{
\scalebox{0.2}[0.25]{\includegraphics{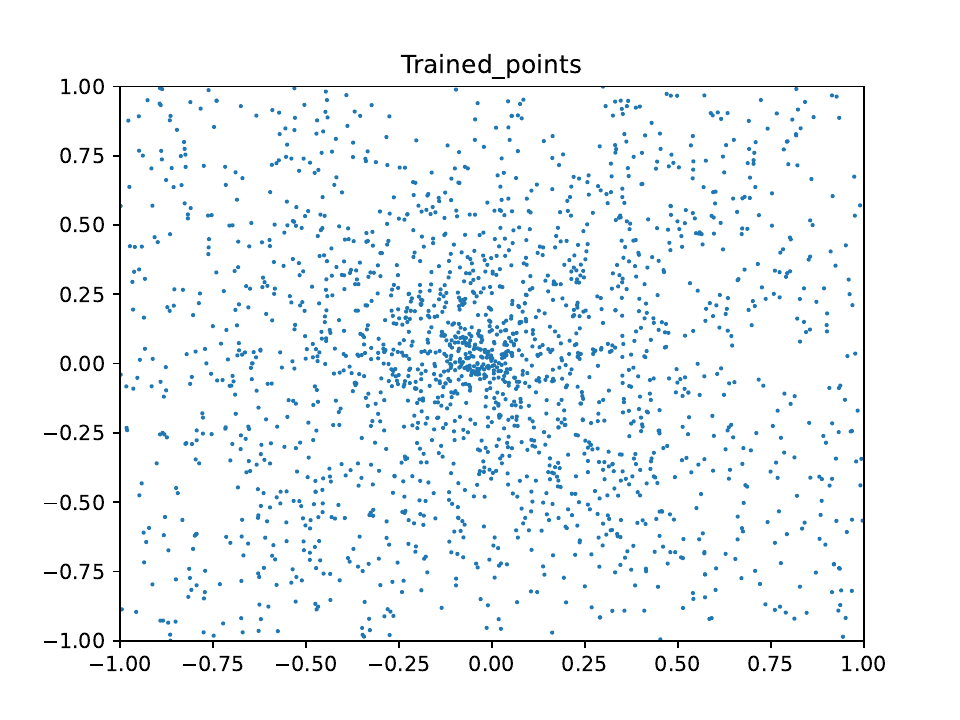}}}

\subfigure[Iteration $6$]{
\scalebox{0.2}[0.25]{\includegraphics{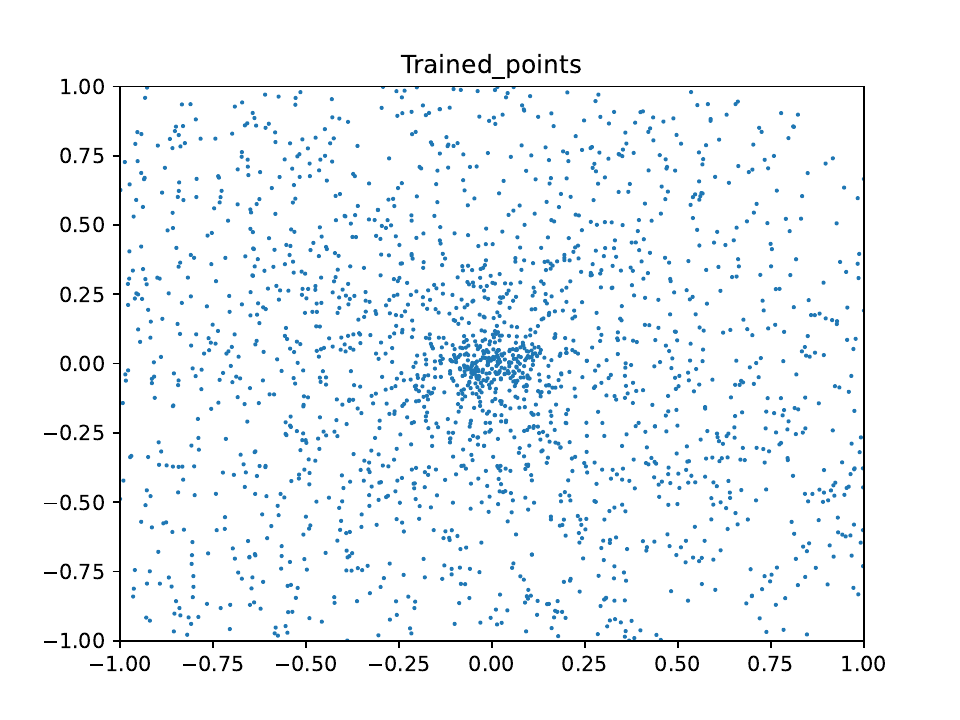}}}
\hspace{5mm}
\subfigure[Iteration $8$]{
\scalebox{0.2}[0.25]{\includegraphics{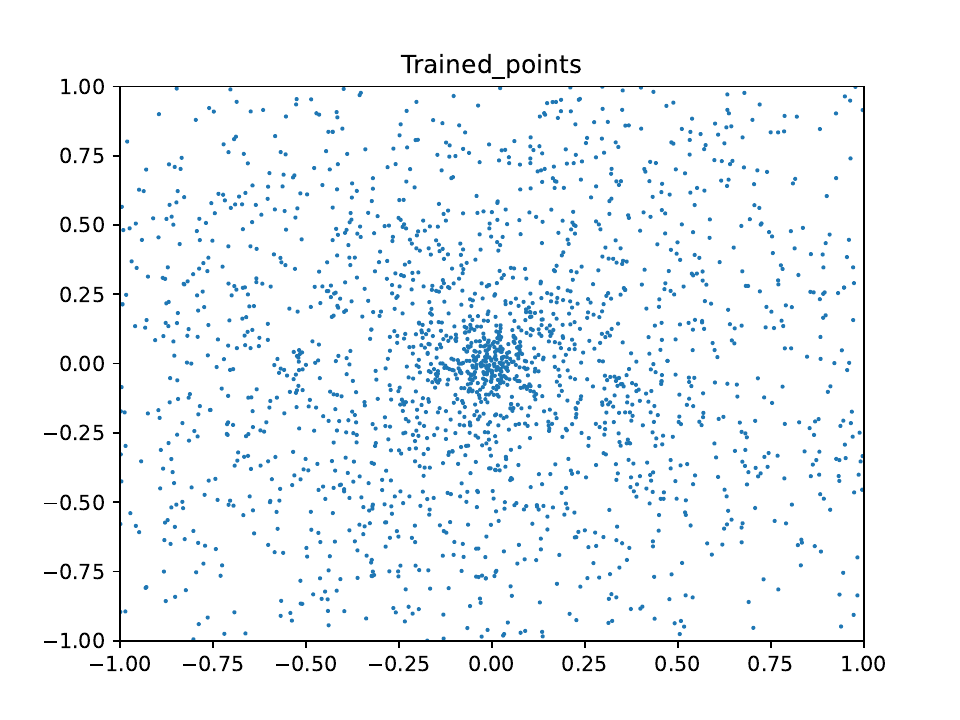}}}
\end{center}
\caption{The evolution of the training points generated from the bounded KRnet.}\label{training points singular}
\end{figure}

\begin{figure}[H]
\begin{center}
\subfigure[Iteration $0$]{
\scalebox{0.3}[0.3]{\includegraphics{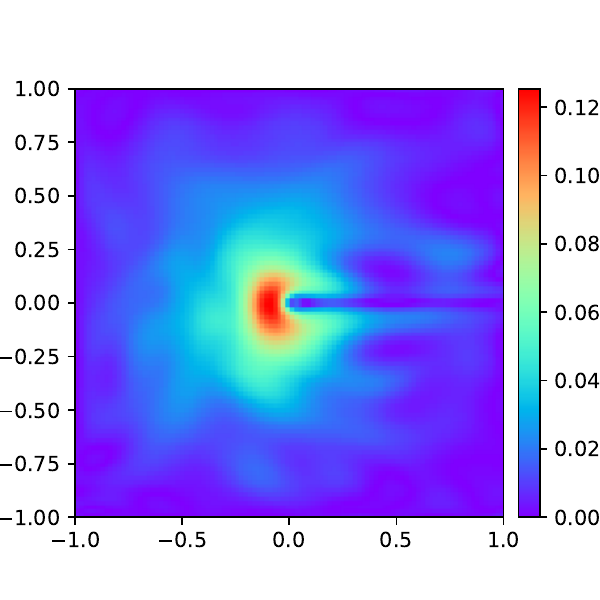}}}\hspace{10mm}
\subfigure[Iteration $2$]{
\scalebox{0.3}[0.3]{\includegraphics{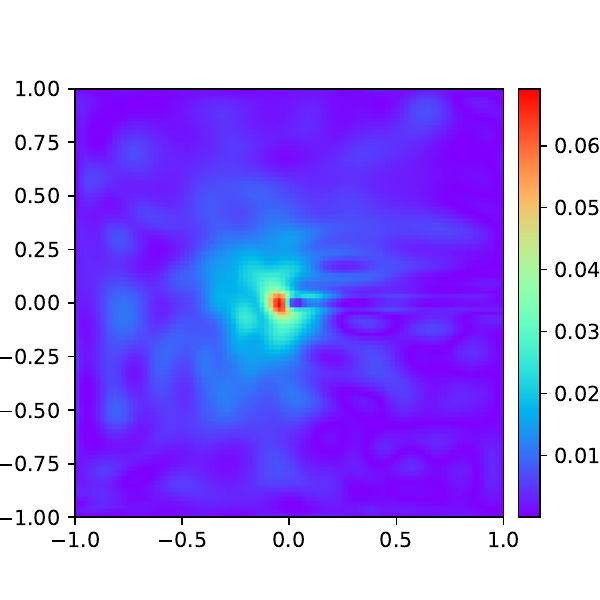}}}
\hspace{10mm}
\subfigure[Iteration $4$]{
\scalebox{0.3}[0.3]{\includegraphics{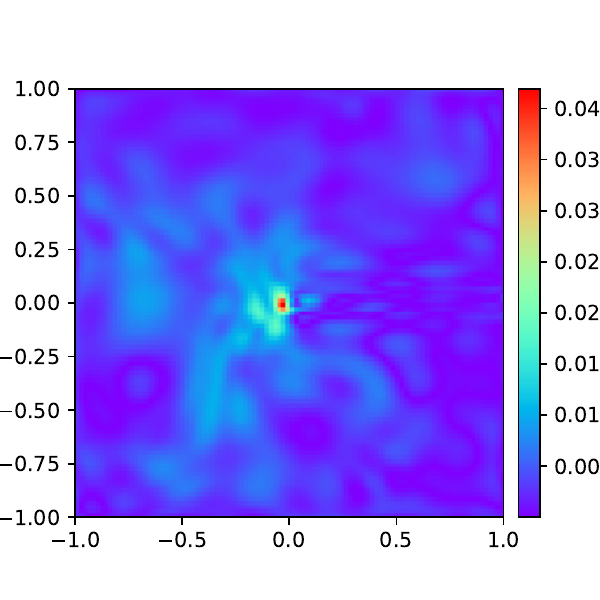}}}

\subfigure[Iteration $6$]{
\scalebox{0.3}[0.3]{\includegraphics{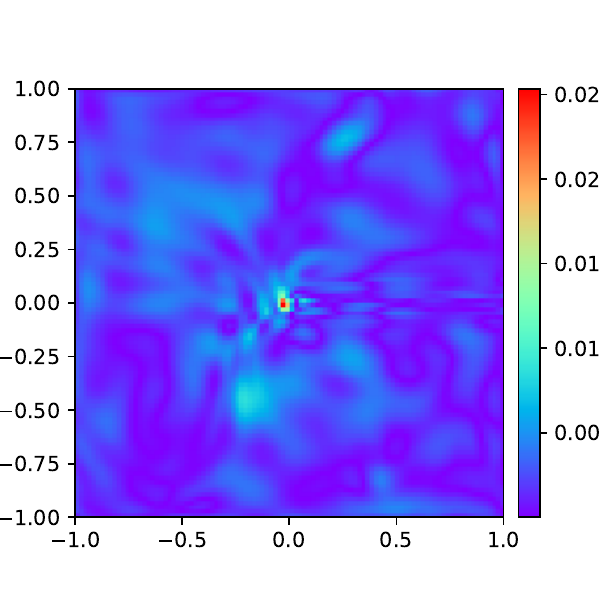}}}
\hspace{10mm}
\subfigure[Iteration $8$]{
\scalebox{0.3}[0.3]{\includegraphics{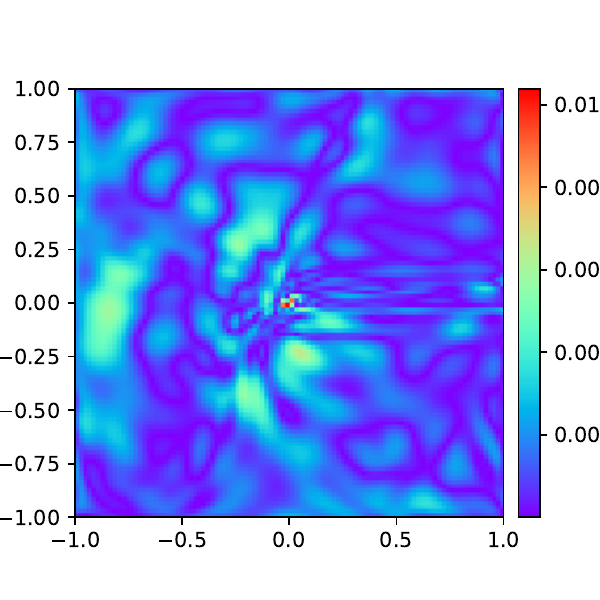}}}
\end{center}
\caption{The evolution of the absolute errors.}\label{abs_err_singular}
\end{figure}

We present a set of numerical results using the pure KRnet sampling approach with a large training set containing $200000$ points. In Figure $\ref{Normalized integrand}$, we display the evolution of the normalized integrand during the training process. It can be observed that the peak (where the singularity occurs) becomes sharper as the number of iterations increases. Since the bounded KRnet  learns a distribution from the normalized integrand, the samples generated from the bounded KRnet become increasingly concentrated around the central peak as shown in Figure $\ref{training points singular}$. In contrast to the problem with a low regularity solution, our adaptive method recognizes that sampling from the region around the singularity becomes increasingly important as the number of iterations grows. Figure $\ref{abs_err_singular}$ illustrates the evolution of the absolute error during the training process. The absolute error around the singularity decreases as the iterations progress.

\begin{figure}[H]
\begin{center}
\subfigure[Uniform]{
\scalebox{0.25}[0.25]{\includegraphics{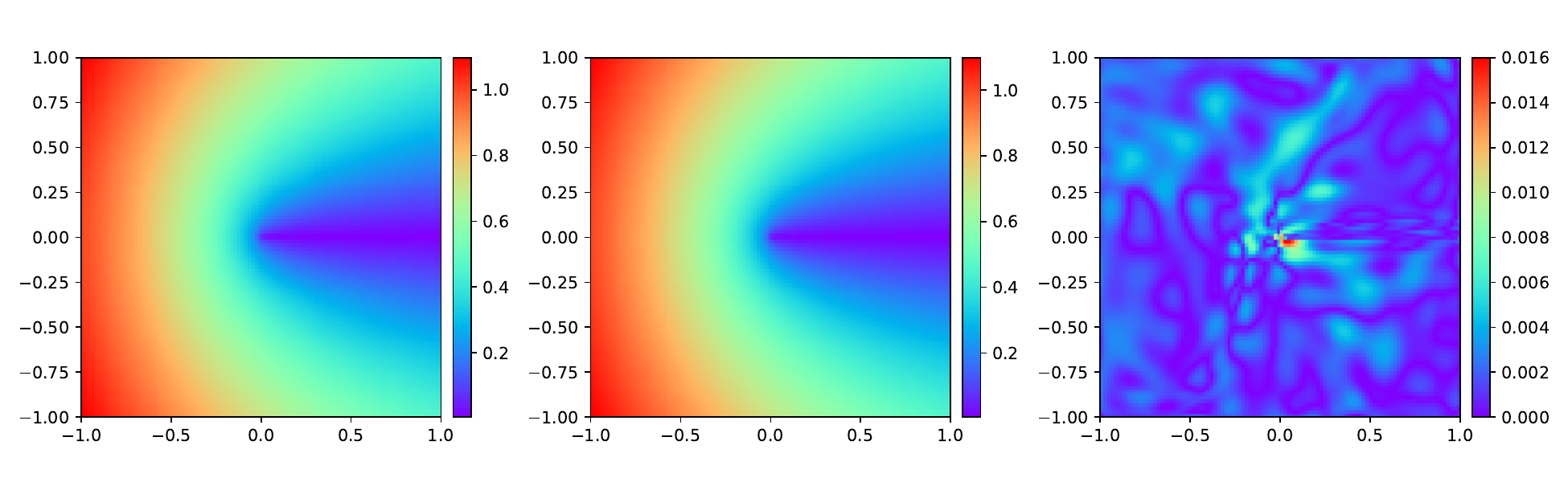}}}
\subfigure[Alternative model $1$ with $\epsilon=0.5$]{
\scalebox{0.25}[0.25]{\includegraphics{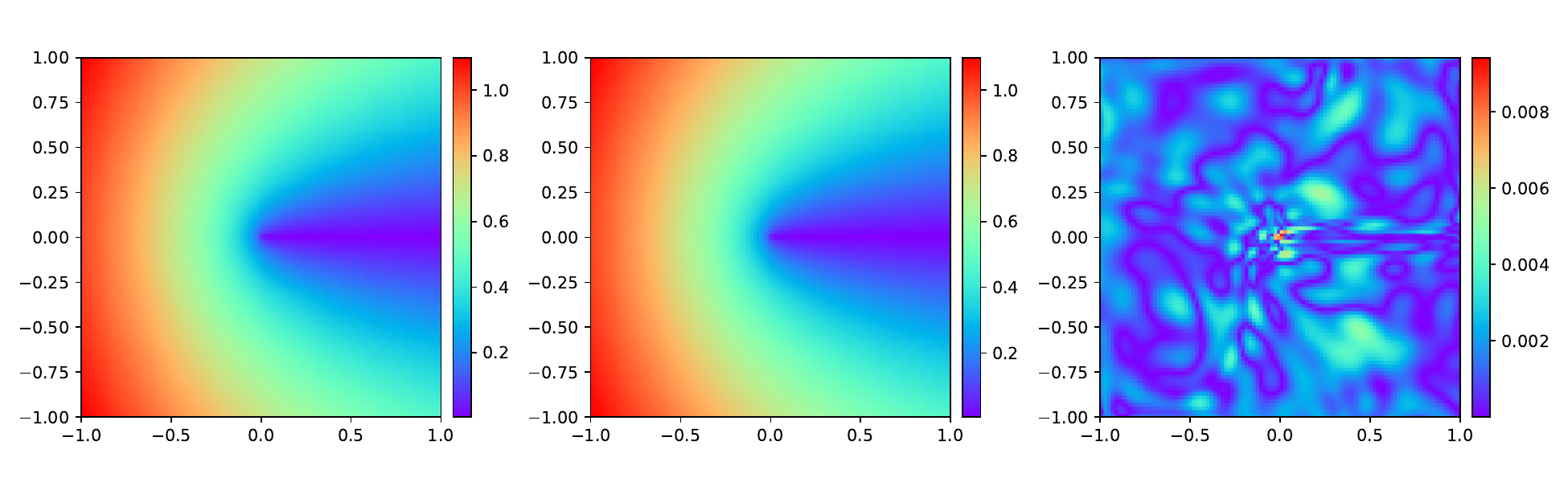}}}
\subfigure[Alternative model $2$ with $\epsilon=0.1$]{
\scalebox{0.25}[0.25]{\includegraphics{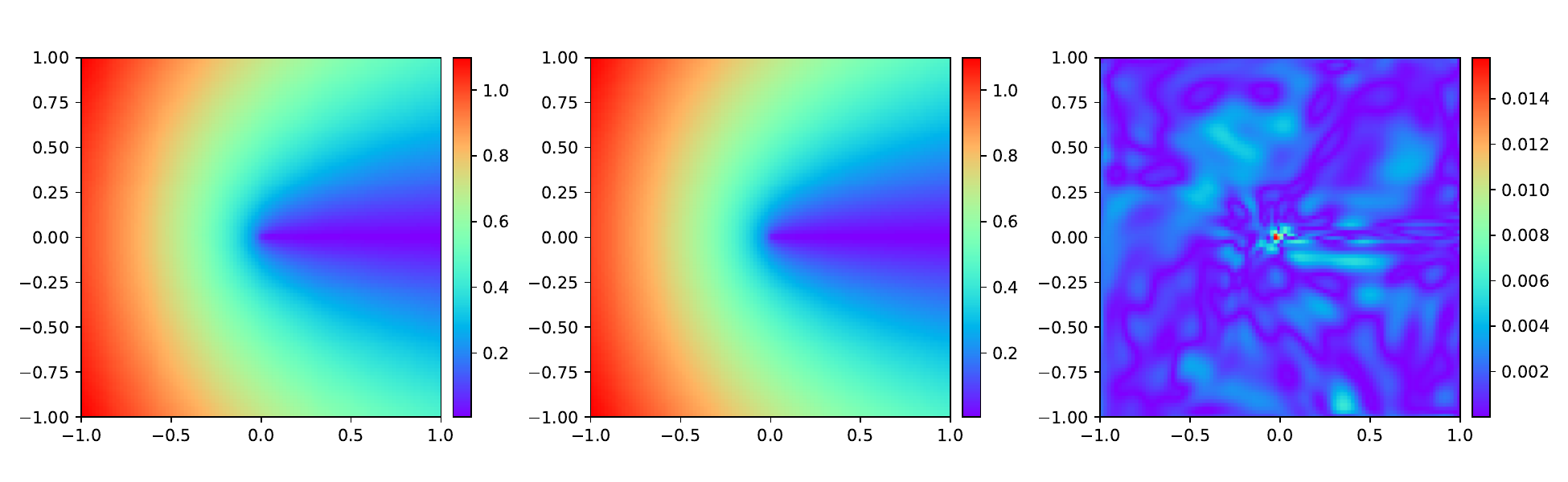}}}
\subfigure[Pure bounded KRnet]{
\scalebox{0.25}[0.25]{\includegraphics{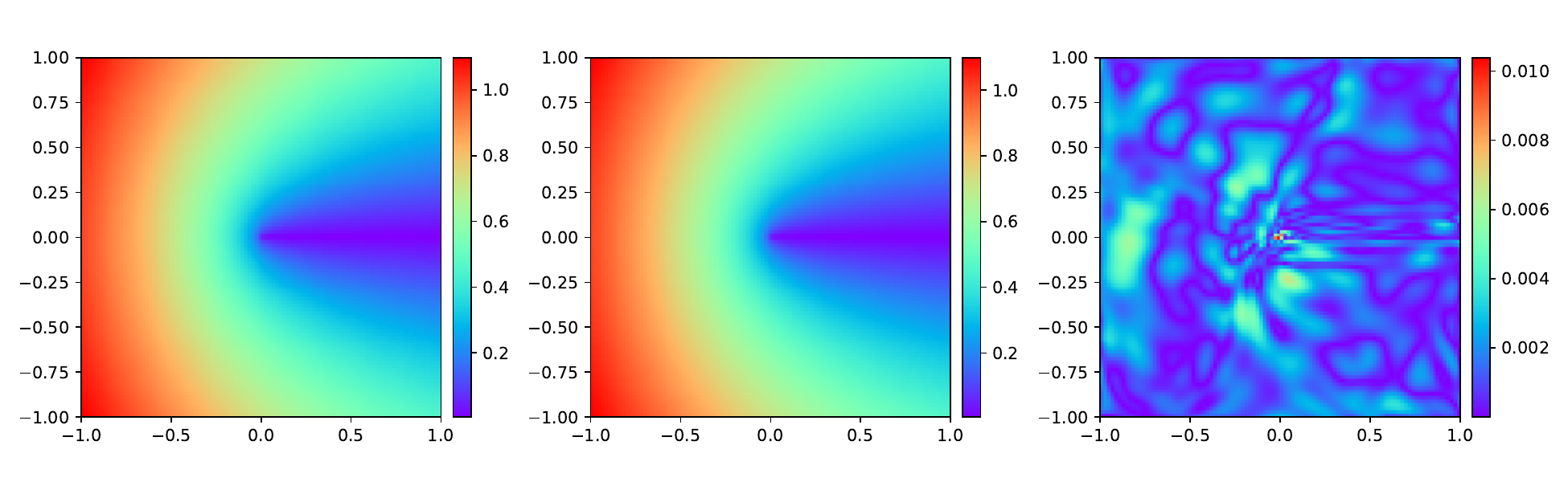}}}
\end{center}
\caption{Comparison of three adaptive sampling methods. From left to right: numerical solution, exact solution, absolute error}\label{Comparison of three adaptive sampling methods}
\end{figure}

\begin{figure}[H]
\begin{center}
\scalebox{0.6}[0.5]{\includegraphics{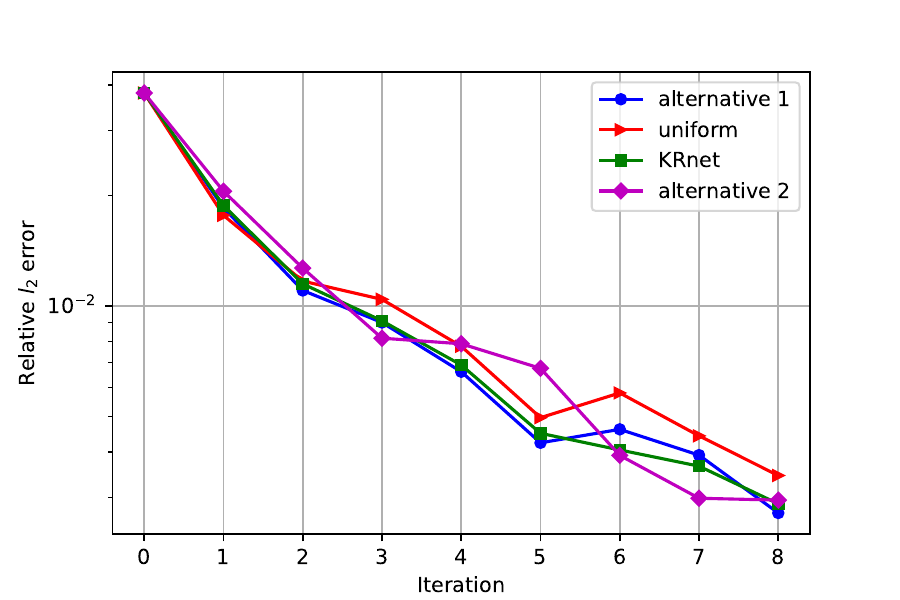}}\end{center}
\caption{Comparison of $L_2$ errors}\label{comparison l2 singular}
\end{figure}

We compare the absolute errors and $L_2$ errors of the four adaptive sampling methods in Figure $\ref{Comparison of three adaptive sampling methods}$ and Figure $\ref{comparison l2 singular}$. Unlike the problem with a low regularity solution, the uniform samples based method works well for this problem. However, its performance is not as good as the other three methods based on deep generative models. The absolute error of the uniform method is almost twice as large as the error of the alternative model $1$ based method. From iteration $3$, the errors of alternative model $1$ based method and pure bounded KRnet based method drop faster than that of the uniform samples based method.

\subsection{High dimensional Poisson problem}
We consider the following $10$D elliptic equation
$$
\begin{aligned}
-\Delta u(x)&=f(x), &x\in \Omega=[-1,1]^{10}
\end{aligned}
$$ 
with an exact solution
$$
u(x)=e^{-10\Vert x\Vert_2^2}.
$$
where the Dirichlet boundary condition on $\partial \Omega$ is given by the exact solution.  In a high-dimensional hypercube, a large portion of uniformly distributed samples tend to cluster near the surface of the hypercube. However, from the exact solution of this 10D Poisson problem, we know the important computation region is mainly in the neighborhood of the origin. Hence, if we employ uniformly distributed samples to train the neural networks, the majority of these samples will contribute little to error reduction.

\begin{figure}[H]
\begin{center}
\subfigure[Calculated solution, True solution, Absolute error]{
\scalebox{0.35}[0.35]{\includegraphics{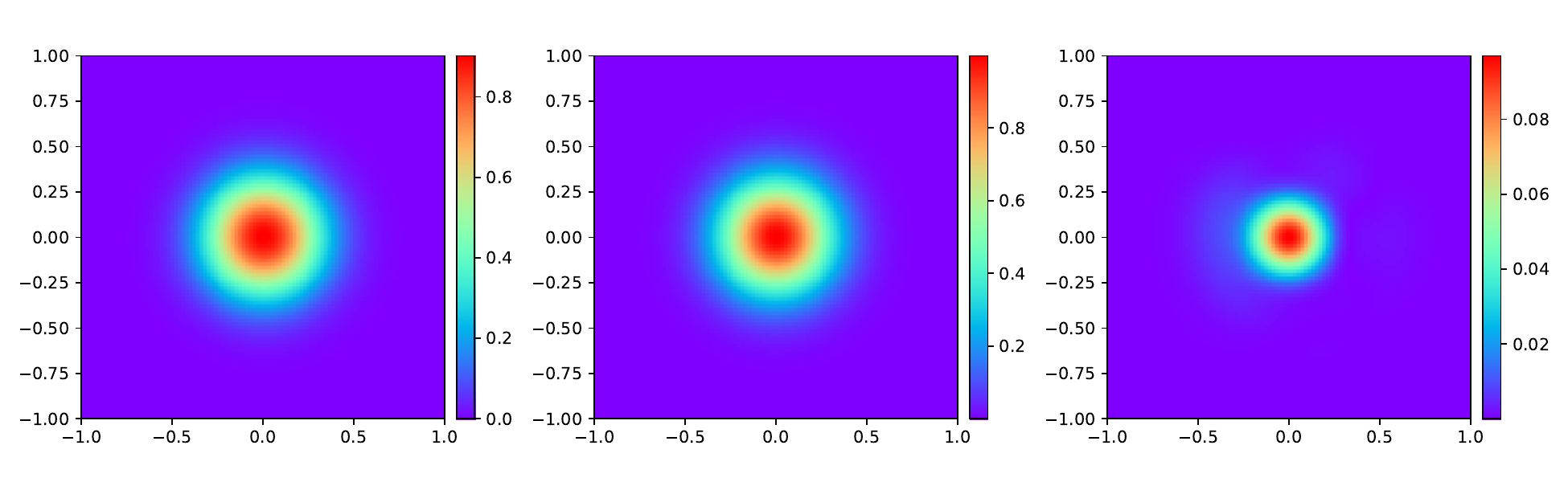}}
}
\end{center}
\caption{Numerical result: Calculated solution, True solution, Absolute error on first two dimension at iteration $7$}\label{result_hd}
\end{figure}

We use a ResNet $u(x,\theta)$ with a stack of $4$ blocks(eight fully-connected layers) and $64$ neurons in each layer. For bounded KRnet, we use $8$ CDF layers and two fully connected layers with $32$ neurons for each CDF layers.  We conduct a total of $7$ adaptive iterations for this problem. The number of epochs for training bounded KRnet is set to $2000$, while for training $u(x,\theta)$, we initially use $10000$ epochs for the first iteration, followed by an additional $1000$ epochs for each subsequent iteration. The balance parameter $\beta$ in the Deep Ritz method is selected to be $1000$. The learning rate for ADAM operator is set to $0.001$ and the batch size is set to $m=20000$.  We apply the alternative PDF model $1$ with $\epsilon=0.5$. We conduct our tests on a test set with $1000000$ points sampled from a uniform distribution on $\Omega$.

We present the numerical solution and the absolute error in Figure $\ref{result_hd}$.  We illustrate the evolution of training points for iterations $1$,$3$, $5$, and $7$, focusing on the components $x_1$ and $x_2$ for visualization purposes. We also examined the other components and did not observe significantly different results. Based on the numerical results and the distribution of training points, our adaptive method effectively captures the crucial computation region around the origin and therefore reduces the error peak.

\begin{figure}[H]
\begin{center}
\subfigure[Training points on first two dimension at iteration $1$]{
\scalebox{0.25}[0.3125]{\includegraphics{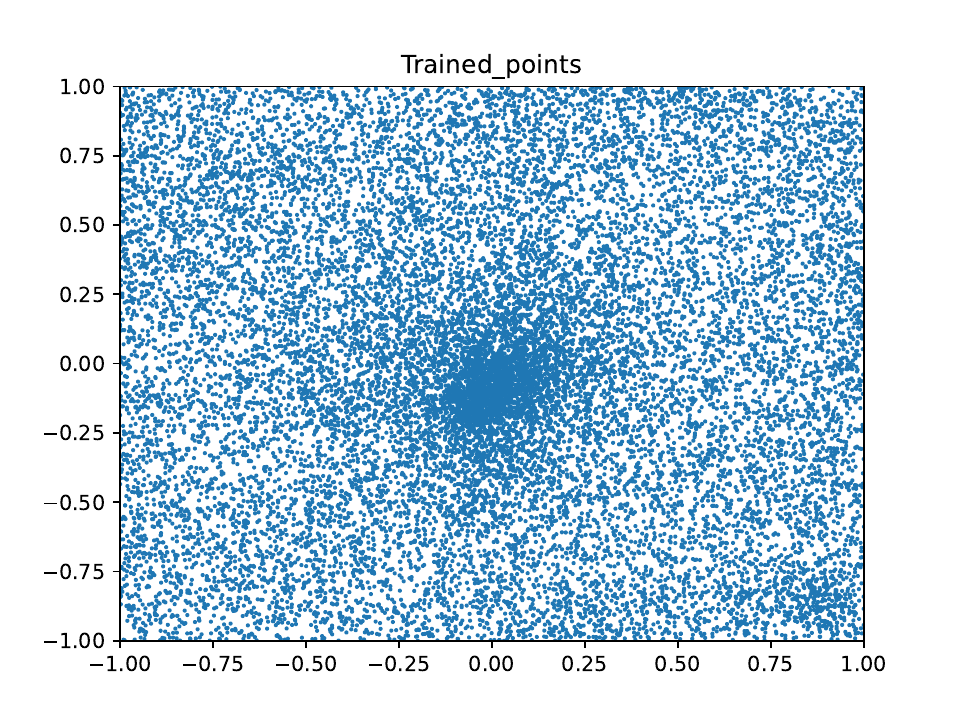}}
}
\hspace{10mm}
\subfigure[Training points on first two dimension at iteration $3$]{
\scalebox{0.25}[0.3125]{\includegraphics{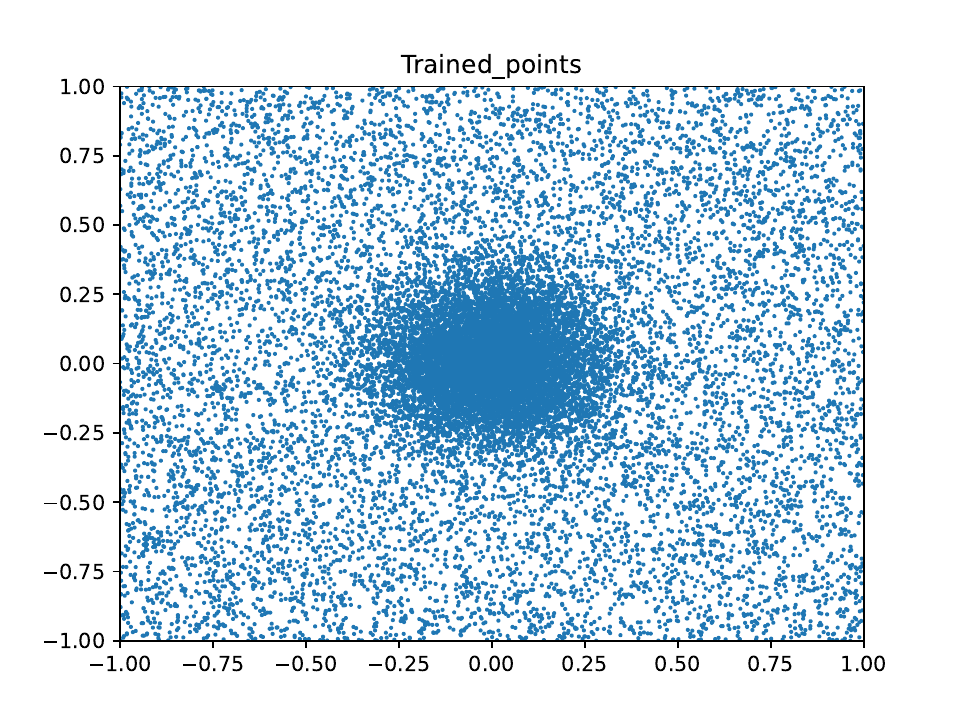}}
}

\subfigure[Training points on first two dimension at iteration $5$]{
\scalebox{0.25}[0.3125]{\includegraphics{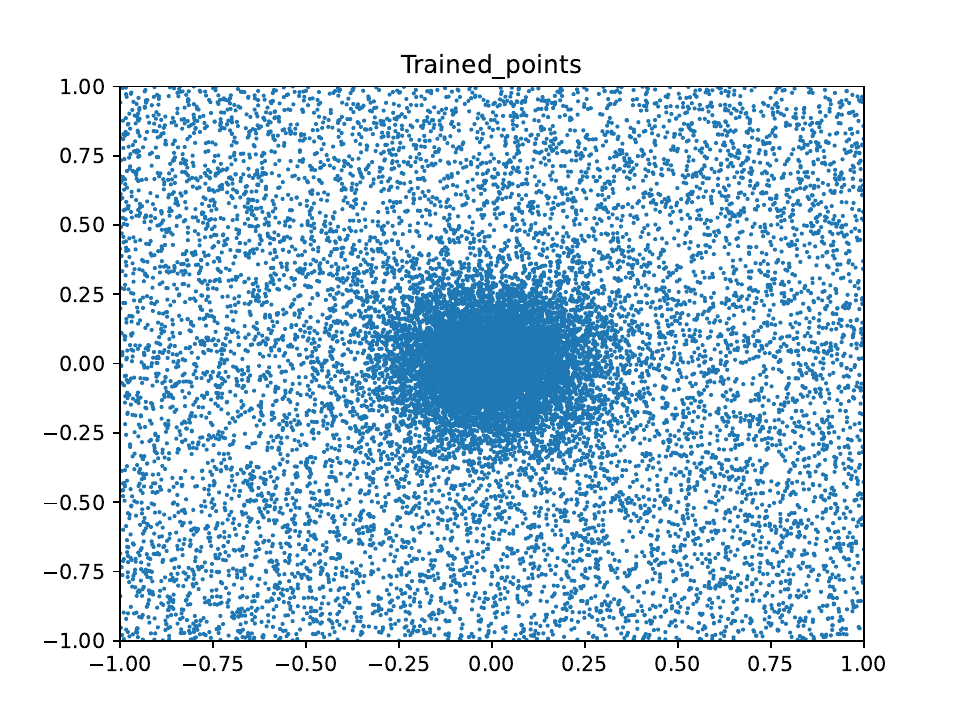}}
}
\hspace{10mm}
\subfigure[Training points on first two dimension at iteration $7$]{
\scalebox{0.25}[0.3125]{\includegraphics{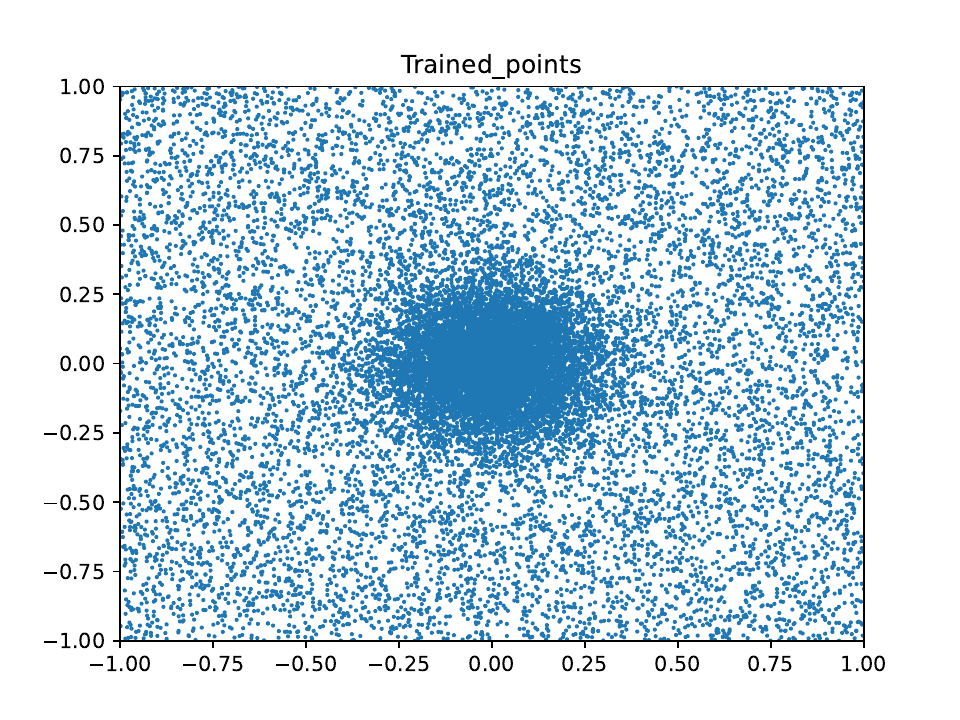}}
}
\end{center}
\caption{The evolution of training points on first two dimensions}\label{train_points_hd}
\end{figure}

\begin{figure}[H]
\begin{center}
\scalebox{0.5}[0.4]{\includegraphics{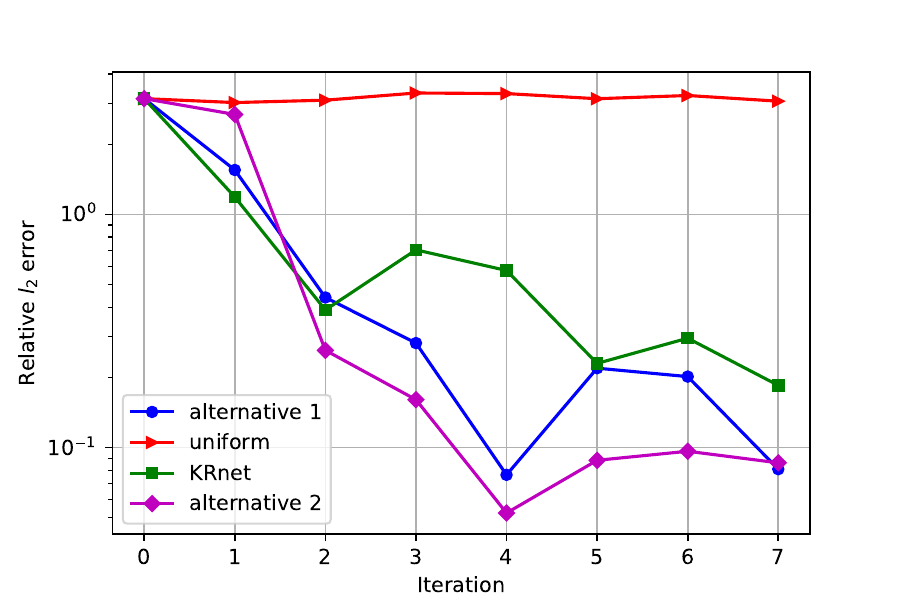}}
\end{center}
\caption{Relative $L_2$ errors}\label{error_hd}
\end{figure}

We compare the $L_2$ errors of four different adaptive sampling strategies in Figure $\ref{error_hd}$. It can be seen that the $L_2$ error of the uniform sampling strategy does not decrease as the iteration progresses. The $L_2$ error of the other three adaptive methods exhibits decreasing trends. Among them, the alternative method $2$ with $\epsilon=0.1$ performs best. It achieves the smallest relative $L_2$ error $5.2451e-02$ at iteration $4$. Both alternative model based methods can be terminated earlier at iteration $4$ to reduce computational cost.

\section{Conclusion}\label{conclusion}
In this work, we propose an adaptive sampling strategy for the Deep Ritz method. We use a deep neural network to approximate the solution of the PDE and a deep generative model called bounded KRnet to approximate the integrand in the variational loss of the Deep Ritz method. Next, we employ the bounded KRnet to generate new samples and calculate the associated probability density function (PDF) values for these samples. The variational loss can then be more accurately calculated using importance sampling and therefore a more precise solution can be obtained from the Deep Ritz method. We present four numerical experiments to illustrate the performance of our adaptive sampling strategy. These numerical experiments include two problems with low regularity, a problem with singularity and a high dimensional problem. In all four experiments, our adaptive sampling strategy consistently achieves a more accurate solution compared to the uniform sampling strategy used in the original Deep Ritz method.  We also develop two alternative PDF models based on bounded KRnet. In our experiments, these alternative PDF models can outperform the PDF directly from a bounded KRnet. There are several possible ways to further improve our adaptive sampling strategy based on bounded KRnet. 
First,  in this work, our adaptive sampling strategies require to replace entire training set at each adaptive iteration. The disadvantage of this approach is that the relative $L_2$ errors do not monotonically decrease. For PINNs, gradually adding training points at each iteration is an effective way to mitigate $L_2$ error oscillations caused by the randomness associated with replacing entire training set~\cite{tang2023pinns}. Therefore, one potential approach to improve current adaptive strategies is to develop a suitable adding points strategy for Deep Ritz method. 
Second, other variance reduction techniques such as control variates can be taken into consideration to further improve the approximation of the variational loss. Research on these issues will be reported in future studies.

\section{Conflict of interest statement}
On behalf of all authors, the corresponding author states that there is no conflict of interest.

\printbibliography
\end{document}